\documentclass{article}

\usepackage{PRIMEarxiv}

\usepackage[utf8]{inputenc} 
\usepackage[T1]{fontenc}    
\usepackage{hyperref}       
\usepackage{url}            
\usepackage{booktabs}       
\usepackage{amsfonts}       
\usepackage{nicefrac}       
\usepackage{microtype}      
\usepackage{lipsum}
\usepackage{fancyhdr}       
\usepackage{graphicx}       
\graphicspath{{media/}}     
\usepackage{subcaption}
\usepackage{multirow}
\usepackage{fancyhdr}
\fancypagestyle{alim}{\fancyhf{}\fancyfoot[L]{* Work done while employed at Bosch Research}}
\title{Domain Adaptation of Synthetic Driving Datasets for Real-World Autonomous Driving
}

\author{
  Koustav Mullick \\
  Robert Bosch Corporate Research \\
  Bangalore, India\\
  \texttt{koustav.mullick@in.bosch.com} \\
   \And
  Harshil Jain* \\
  KLA \\
  Chennai, India\\ 
   \AND
  Sanchit Gupta* \\
  Birla Institute of Technology and Science, Pilani \\
  Hyderabad, India \\
   \And
  Amit Arvind Kale \\
  Robert Bosch Corporate Research \\
  Bangalore, India\\
  \texttt{amitarvind.kale@in.bosch.com} \\
}

\begin{document}
\maketitle

\thispagestyle{alim}

\begin{abstract}
While developing perception based deep learning models, the benefit of synthetic data is enormous. However, performance of networks trained with synthetic data for certain computer vision tasks degrade significantly when tested on real world data due to the domain gap between them. One of the popular solutions in bridging this gap between synthetic and actual world data is to frame it as a domain adaptation task. In this paper, we propose and evaluate novel ways for the betterment of such approaches. In particular we build upon the method of UNIT-GAN \cite{unitgan}.

In normal GAN training for the task of domain translation, pairing of images from both the domains (\textit{viz, real and synthetic}) is done randomly. We propose a novel method to efficiently incorporate semantic supervision into this pair selection, which helps in boosting the performance of the model along with improving the visual quality of such transformed images. We illustrate our empirical findings on Cityscapes \cite{cityscapes} and challenging synthetic dataset Synscapes \cite{synscapes}. Though the findings are reported on the base network of UNIT-GAN, they can be easily extended to any other similar network.
\end{abstract}


\section{Introduction}
Autonomous Driving is one of the fastest-evolving, cutting-edge research field currently, with a massive market potential in the future \cite{admarket}. A typical autonomous vehicle is equipped with different types of sensors, such as Cameras, LiDARs, and RADARs. In this regards, one of the key task involves training machine-learning models on the vast amount of data collected by the fleets of cars fitted with these sensors, to assist autonomous driving. But there is a limitation on the size and variety of scenarios that can be captured from the real world with a finite amount of cars moving around. To address this problem, an alternative is to use Synthetic data to train the machine learning algorithms. For example, Waymo has tested its autonomous vehicles by driving only 8 million miles on real roads, compared to 5 billion miles on simulated roadways, as reported by their CEO \cite{waymo}.

\begin{figure}[t]
	\centering
	\begin{subfigure}{0.32\linewidth}
		\centering
		\includegraphics[width=\linewidth]{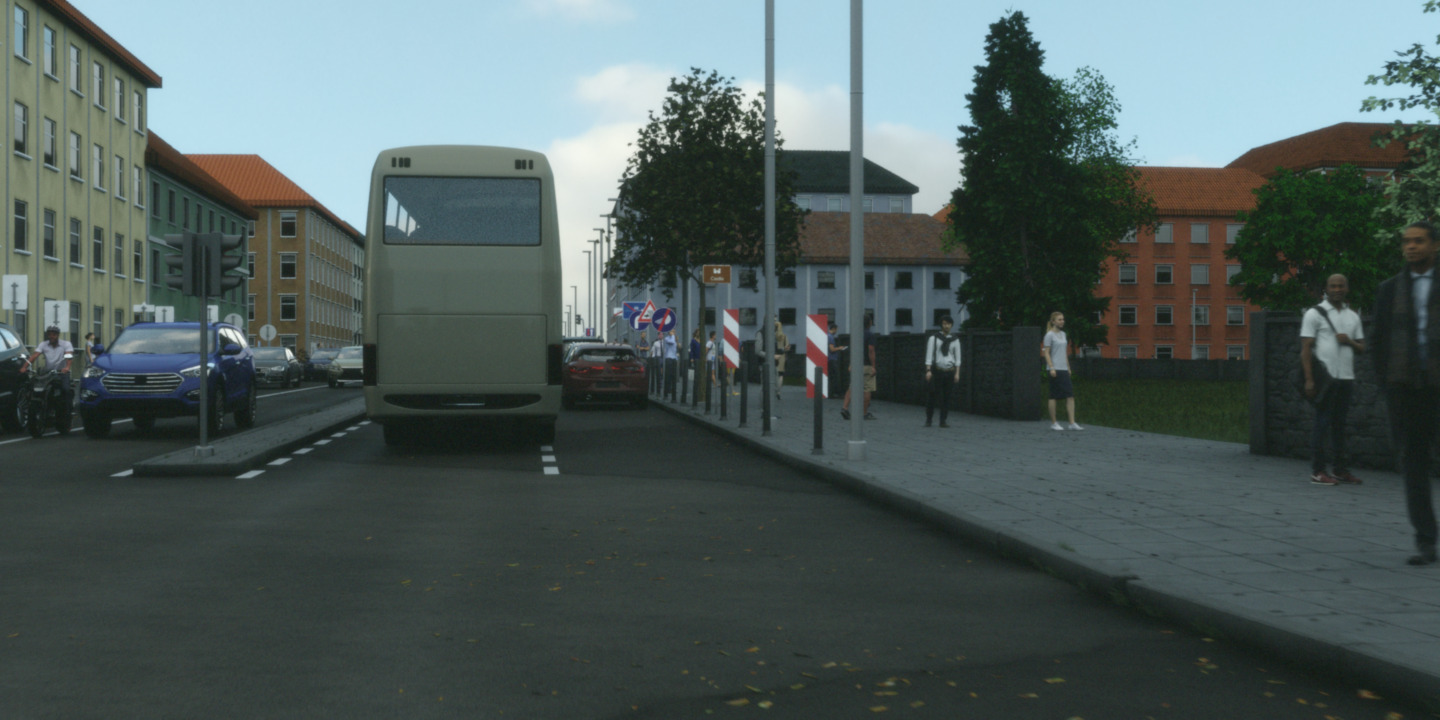}
		\caption{\small Original Synscapes}    
	\end{subfigure}
	\begin{subfigure}{0.32\linewidth}  
		\centering 
		\includegraphics[width=\linewidth]{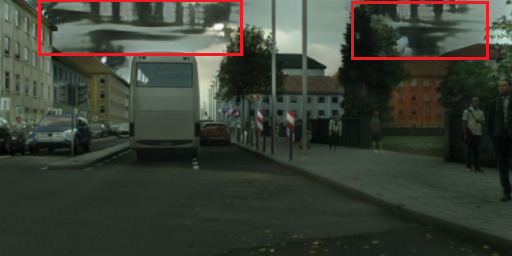}
		\caption{\small UNIT-GAN output}
	\end{subfigure}
	\begin{subfigure}{0.32\linewidth}   
		\centering 
		\includegraphics[width=\linewidth]{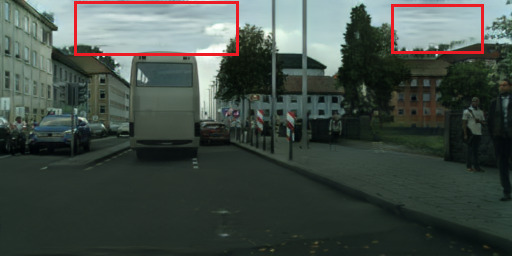}
		\caption{\small Our output}
	\end{subfigure}
	\caption{Overview image showing generated photorealistic outputs using different methods on a Synscapes image. \textit{Red boxes} highlight arbitrary noises induced by actual UNIT-GAN (b), which are not there in our proposed SP-Feat based approach (c). Best viewed in color and by zooming in.}
	\label{fig:intro}
\end{figure}

Learning from synthetic data, where labels are readily available, sounds promising. However, the performance of networks trained with synthetic data for certain computer-vision tasks degrade significantly when applied on novel realistic data. This is mainly due to domain discrepancies \cite{fcn} between them. We explore a method that reduces the domain gap between real world and synthetic data distribution by framing it as a domain adaptation task using Generative Adversarial Networks \cite{gan}. We show how incorporating semantic resemblance across the two domains while training such a GAN architecture can give significant improvement. The proposed approach is based on a simple, robust feature derived from the semantic layout of the images, which we call as the \textit{Spatial Pyramid Features (SP-Feat)}. This feature is incorporated in the batch selection strategy for training generative adversarial network. Both quantitative and qualitative results prove its effectiveness. While, as another widely used alternative, we considered benchmarking with style-transfer based approaches \cite{closed_form_soln} to render a target image in the style of a source image, we observe it works best only when provided with semantic segmentation supervision during inferencing, which is an onerous requirement for usability. In contrast our method only requires such pairing during training, not during inference.

In figure \ref{fig:intro} we present output generated using our approach, over baseline method. The significance of the highlighted regions are explained further into the paper. Further, we try to derive how achieving better performance after inducing photorealism on related computer vision task is correlated to the perceptual quality of the photorealistically rendered synthetic image. Towards this, we carry out extensive evaluations using appropriate validation techniques, both quantitative and qualitative, to show the contribution of the proposed approach. We use Cityscapes \cite{cityscapes} as representative of real world images, and GTA-V \cite{gtav}, Synscapes \cite{synscapes} for synthetic images.  However, we carry out most of the detailed analysis on the Synscapes dataset, and not on GTA-V. This is because the quality of Synscapes is far better than GTA-V. Thus improving it further is more challenging and valuable. 

Towards making synthetic data more usable for real world consumption in autonomous driving paradigm, the contributions of this paper are: \textit{i) We propose a smart batch selection technique to improve domain adaptation task using GANs. ii) Demonstrate the advantage of using such an approach in generation of visually better quality translated images from synthetic to real-world domain. iii) Explore how such an improvement also correlates to better feature adaptation between the two domains, ultimately leading to better performance on downstream computer vision tasks. iv) Finally, this is the first work, to the best of our knowledge, to carry out substantial qualitative and quantitative evaluation on the Synscapes dataset \cite{synscapes}.}

\section{Related Work}

The idea of image translation between domains goes back to Hertzman \textit{et al} \cite{image_analogies}, who employed a non-parametric texture model \cite{texture_synthesis} on a single input-output training image pair. More recently, neural generative models including generative adversarial networks \cite{gan}, variational autoencoders (VAE) \cite{vae}, attention models \cite{attention_model}, moment matching\cite{moment_matching} , stochastic back-propagations \cite{stochastic_backprop}, and diffusion processes \cite{diffusion_process}, have shown that a deep neural network can learn an image distribution from a given set of samples. Several translation models are built upon such generative networks. Given paired input-output images, Judy \textit{et al.} \cite{fcn} proposed a parametric translation function using convolutional neural networks for image translation. In ``pix2pix” \cite{pix2pix}, Isola \textit{et al.} proposed a conditional generative adversarial network \cite{gan} to learn a mapping between input and output images in a paired setting. Similar ideas have also been applied to tasks such as generating photographs from sketches \cite{sketch_learning} or from attribute and semantic layouts \cite{attribute_learning}. These methods are however, limited because of their paired setting which might not always be the most feasible approach, such as in our case of adopting synthetic driving data to real world.

Li \textit{et al} \cite{closed_form_soln} and Fujun \textit{et al.} \cite{deep_style_transfer} have tried to formulate the image translation problem as style transfer, in which they try to transfer style of reference image to the content image. However, one limitation of the style transfer approach is that during inference the networks has to be presented with both the content and style image for translation. In \cite{learning_to_generate,spade} the photorealistic images are obtained from semantic maps, which are provided as input to the network. Also, \cite{cycada,fcn} exploit semantic maps while training GANs by incorporating a losses in the network explicitly, to boost semantic segmentation performance. However, the focus of these works is to solely improve the performance of semantic segmentation using synthetic data. Some other works such as \cite{infogan,discogan,diverse_translation,DRIT_plus}, try to disentangle the representation into content and style. The domain translation is achieved by combining a particular content with the latent representation of the desired style. Given the complexity of their approach, it is difficult to adapt such a framework for complicated driving scenes. Some methods also tackle domain adaptation in an unpaired setting. Here the goal is to relate two data domains, in the absence of exact one-to-one mapping between them. Ming \textit{et al} \cite{cogan}  and Yusuf \textit{et al.} \cite{cross_modal} learns a domain agnostic, joint shared distribution of multi-domain images. Concurrent to the implemented method, Zhu et al. \cite{cyclegan}, introduced the CycleGAN, depicting the effectiveness of enforcing a cycle-consistency loss in trying to map two distributions of images. In this work, we adopt the UNIT \cite{unitgan} framework, for our well defined task of domain translation from synthetic to real data. UNIT borrows ideas from the previously mentioned works to address the task of domain translation between two domains. MUNIT \cite{munit} extends UNIT to project one source domain across multiple target domains.

\section{Method}

\subsection{Photorealism as Domain Adaptation}

Inspired from the work by Ming-Yu \textit{et al.} \cite{unitgan}, we frame the problem of inducing photorealism in synthetic data as a domain adaptation task between real world and synthetic data. The goal of unpaired image-to-image translation is to learn a joint distribution of images in different domains by using images from the marginal distributions of the individual domains. Since there exists an infinite set of joint distributions that can arrive to the given marginal distributions, one could infer nothing about the joint distribution from the marginal distributions without additional assumptions. We briefly introduce the UNIT-GAN architecture in the next paragraph.

\begin{figure}
	\centering
	\includegraphics[width=0.8\linewidth]{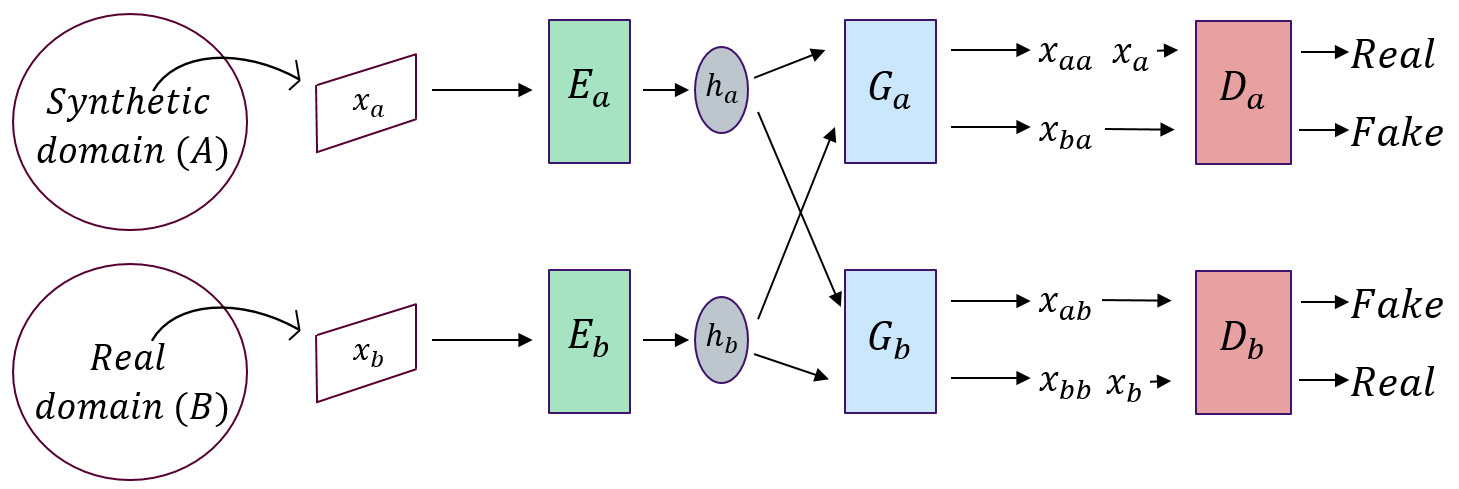}
	\caption{UNIT-GAN Architecture.}
	\label{fig:unit_gan}
\end{figure}

\subsubsection{Unsupervised Image-to-image Translation (UNIT-GAN):}

UNIT makes use of a shared-latent space assumption and devises an unsupervised image-to-image translation framework based on Coupled GANs. Figure \ref{fig:unit_gan} shows the basic block diagram of the UNIT architecture. Given two domains $A$ and $B$, $x_{a,b}$, $E_{a,b}$, $h_{a,b}$, $G_{a,b}$ and $D_{a,b}$ are the \textit{input image, encoder, hidden latent representation, decoder (generator)} and \textit{adversarial discriminator} for $A$ and $B$ respectively. The network is trained using a combination of \textit{reconstruction, cycle-consistency and adversarial losses}. Readers are requested to take a look at the original paper for further details on UNIT-GAN framework.

For convenience, we briefly summarize from the actual paper, the important loss functions related to the training of the network. 

\begin{description}
	\item[$\bullet$] \textit{Reconstruction Loss}: The aim of reconstruction loss is to preserve consistency between the input image and the output we obtain by passing the input image through its own pair of encoder and generator. Ideally, we should get back the same image as the input, from the output of the decoder.
	
	\item[$\bullet$] \textit{Cycle-consistency Loss}: The shared-latent space assumption implies cycle-consistency constraint \cite{cyclegan}, which is included in the framework to further regularize the task.\
	
	\item[$\bullet$] \textit{Adversarial Loss}: To make the generated output for each domain look more realistic, an adversarial loss is included, which differentiates between actual (real) and generated (fake) images.

\end{description}

\subsection{Improved Domain Adaptation}

Figure \ref{fig:unit_gan_output} shows some of the qualitative results obtained on the Synscapes dataset when adapted to Cityscapes style using the above approach. It can be observed that the images have semantic inconsistencies, in terms of noisy patches, across them. The reason for this can be attributed to the batch selection strategy while training the GAN. In the current random sampling strategy, there is no way to enforce semantic consistency across the pairs of images sampled from both the real and synthetic domain. An example scenario can be: During a particular iteration of training, images coming from Synscapes happen to have a lot of pedestrians along the road. Whereas they are not there in the images sampled from Cityscapes. Discriminators of any GAN, are notorious for picking up such subtle differences in order to distinguish between \textit{real} and \textit{fake}. In this context, it plays a very important role while training the network as it makes up a major chunk of the loss flowing through the generator, in order to allow it to generate more realistic looking outputs. In the absence of any additional supervision, the task for the generator becomes very difficult in such a scenario as there is no way for it to make any substantial improvement based on the \textit{weak (or easy) signal} propagated from the discriminator.

\subsubsection{Smart Batch-Loader:}

In order to circumvent the above mentioned possible drawback and prevent the discriminator from picking up on niche queues of semantic differences, it is important to enforce consistency across the image pairs occurring in a particular batch. Semantic inconsistencies across batches of the two domains while training an image translation network could potentially lead to unstable training of the discriminator. We demonstrate this in the next section. We show how rectifying that could be beneficial to obtain better translations, backed by improved qualitative and quantitative results. We propose a modified batch selector based on the semantic labels of the images, which are available to us for the training data. It makes use of a feature derived from the class labels of the images, which we call as the \textit{Spatial Pyramid Features (SP-Feat)}, and is obtained as described in the next section. The requirement of the availability of segmentation ground-truth for this purpose is rather loose. Though we use the available segmentation annotation for finding such images having similar object classes, but it can easily be replaced with coarse annotation using off-the-shelf segmentation network (weak supervision), or even features which work across domains (real and synthetic in our case) to find such pairs. Moreover, during inference no such supervision is needed.

\subsubsection{Spatial Pyramid Features (SP-Feat):}

\begin{figure}
	\centering
	\includegraphics[width=0.65\linewidth]{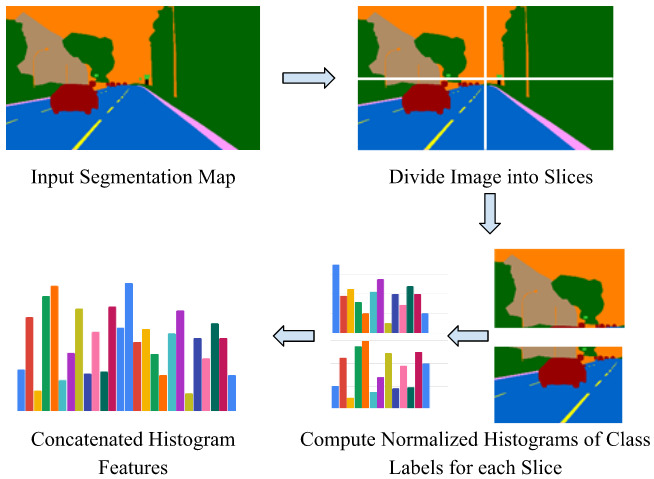}
	\caption{Spatial Pyramid Feature extraction method overview.}
	\label{fig:spfeat}
\end{figure}

For each image from either domain, we consider their corresponding semantic segmentation masks. They can be obtained from the segmentation ground-truth, if available, or can be obtained in a semi-supervised manner by taking predictions from a trained segmentation network. To induce spatial correspondence, we further divide the image into non-overlapping slices, depending on the granularity level we wish to obtain the features at. Then for each slice, we compute histograms of the pixels belonging to each class and concatenate them across all the slices. So the final feature vector is of dimension \textit{(number of classes X number of slices)}. Figure \ref{fig:spfeat} shows an overview of the algorithms. This algorithm is robust, efficient and simple to calculate, enabling it to be scaled easily across tens of thousands of images, without much overhead. The only requirement is that the class structure of the data-sets under consideration must be in agreement.

\begin{figure*}
	\centering  
	\begin{subfigure}{0.9\linewidth}  
		\centering  
		\includegraphics[width=0.32\linewidth]{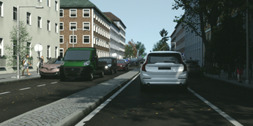}
		\includegraphics[width=0.32\linewidth]{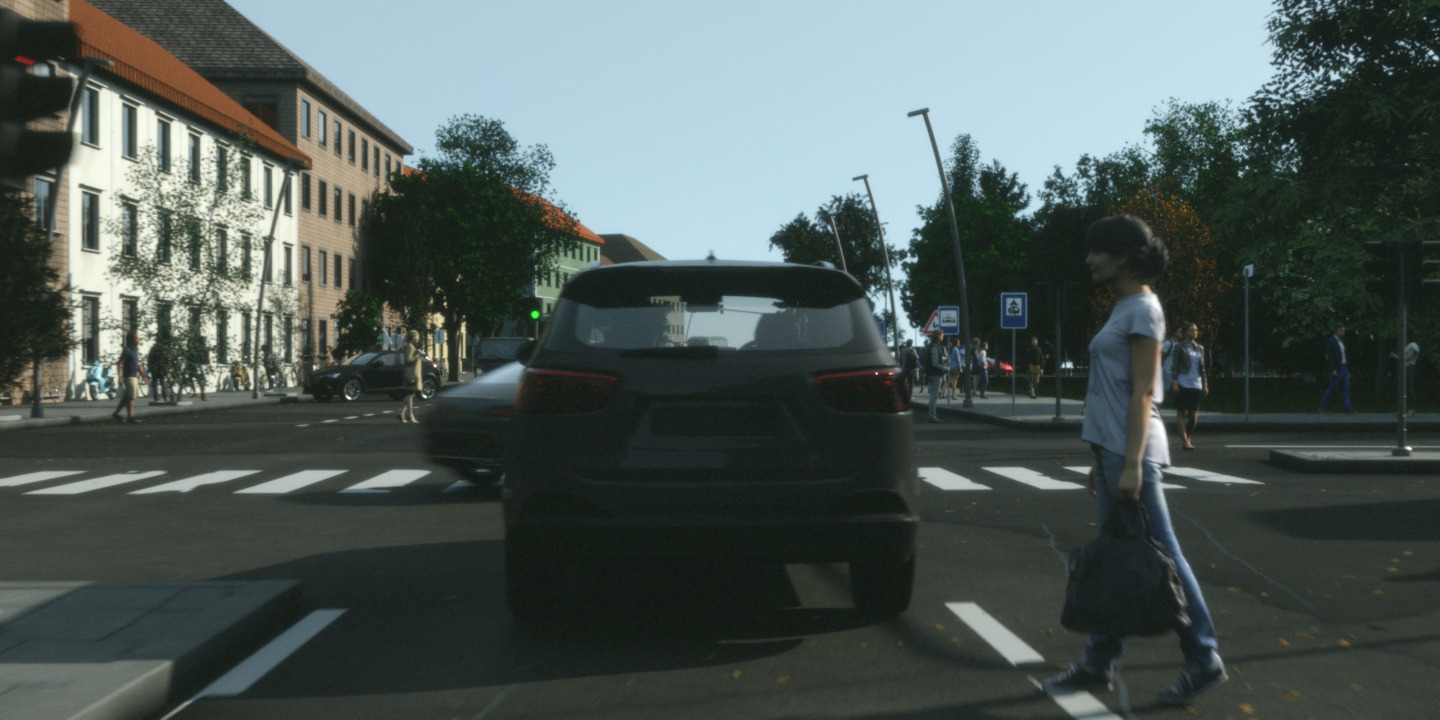}  
		\includegraphics[width=0.32\linewidth]{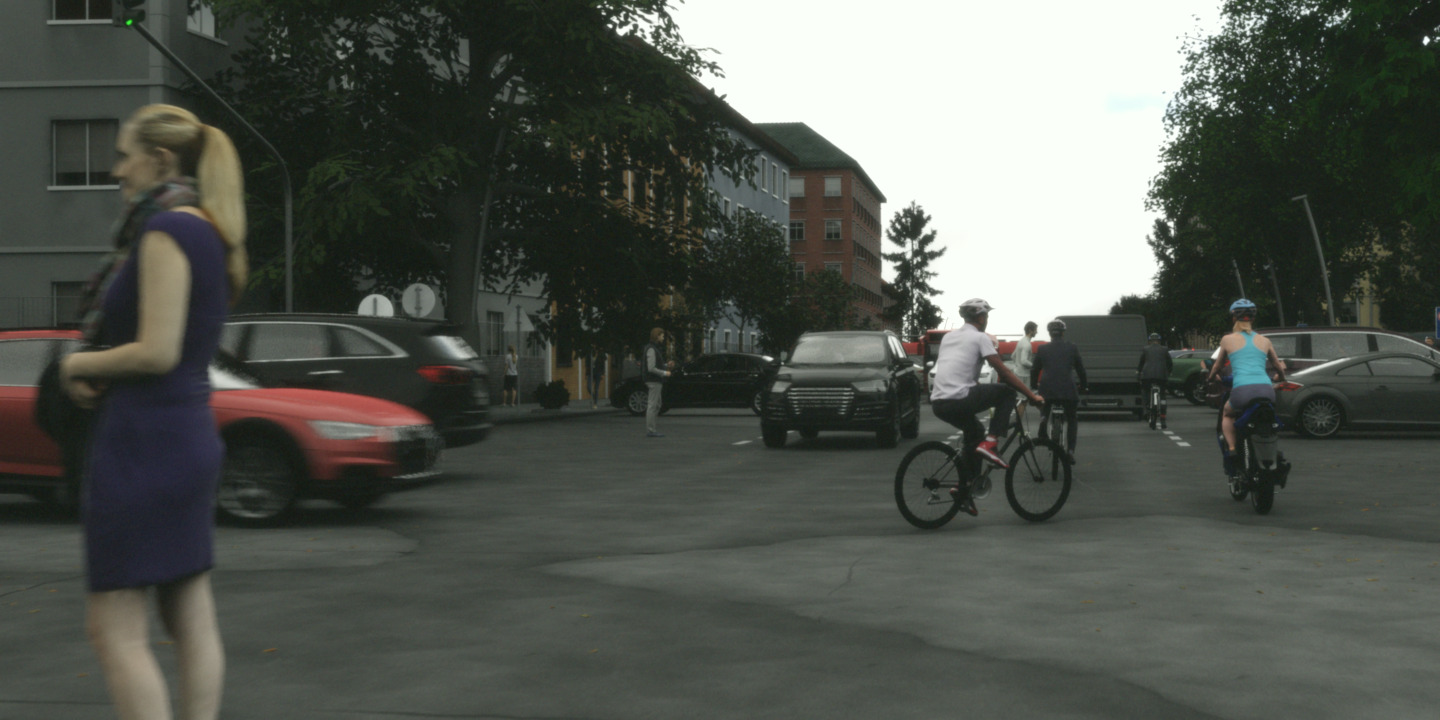}  
		\caption{Example Synthetic images from Synscapes.}
		\label{fig:synscape_input}
	\end{subfigure}   
	
	\begin{subfigure}{0.9\linewidth}  
		\centering  
		\includegraphics[width=0.32\linewidth]{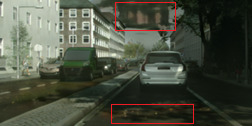}
		\includegraphics[width=0.32\linewidth]{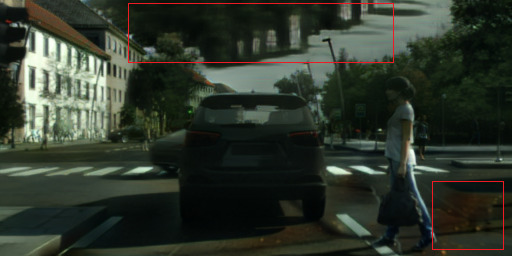}  
		\includegraphics[width=0.32\linewidth]{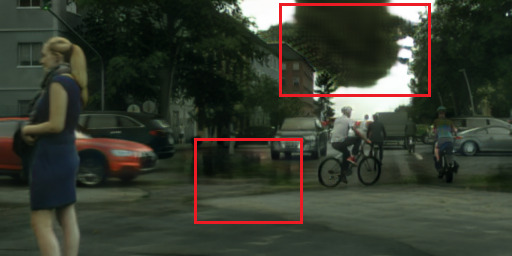}  
		\caption{Results using original UNIT-GAN architecture.}
		\label{fig:unit_gan_output}
	\end{subfigure}
	
	\begin{subfigure}{0.9\linewidth}  
		\centering  
		\includegraphics[width=0.32\linewidth]{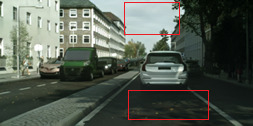}
		\includegraphics[width=0.32\linewidth]{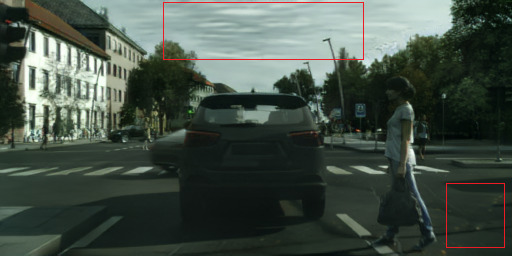}  
		\includegraphics[width=0.32\linewidth]{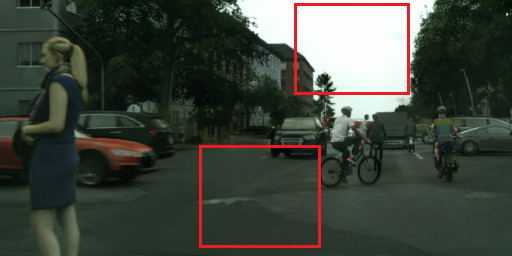}  
		\caption{Results using SP-Feat based Batch selection strategy.}
		\label{fig:spfeat_output}
	\end{subfigure}
	
	\caption{Advantage of Smart Batch-Loader for inducing Photorealism in Synscapes, from Cityscapes. \textit{Red boxes} highlight some regions showing our proposed approach gets rid of noises induced using basic UNIT-GAN network, while inducing photorealism. Best viewed in color.}
	\label{fig:qualitative_result_1}
\end{figure*}

\begin{figure*}
	\centering  
	\begin{subfigure}{0.30\linewidth}  
		\centering  
		\includegraphics[width=\linewidth]{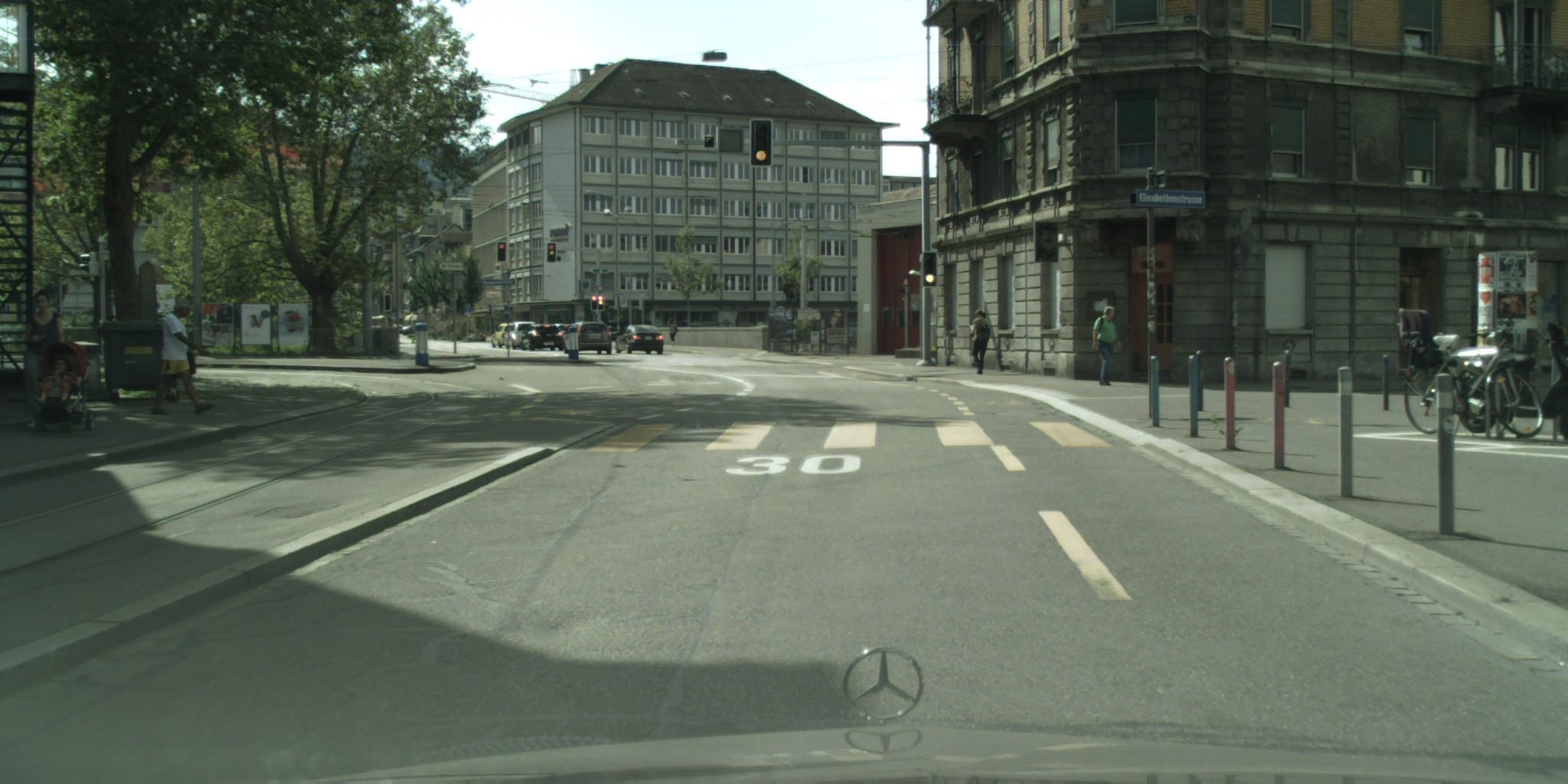}
		
		\vspace{0.2em}
		
		\includegraphics[width=\linewidth]{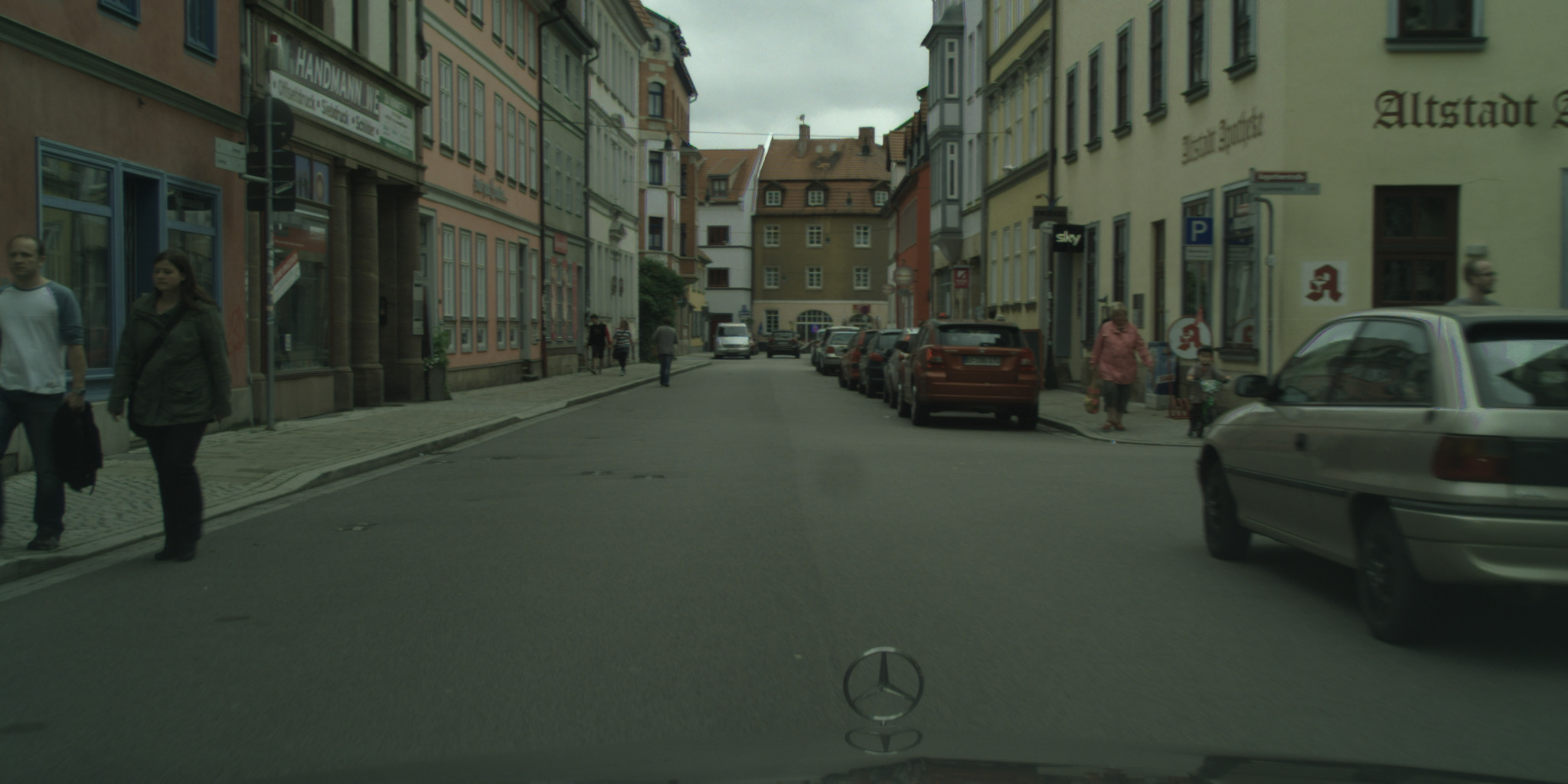}  
		
		\vspace{0.2em}
		
		\includegraphics[width=\linewidth]{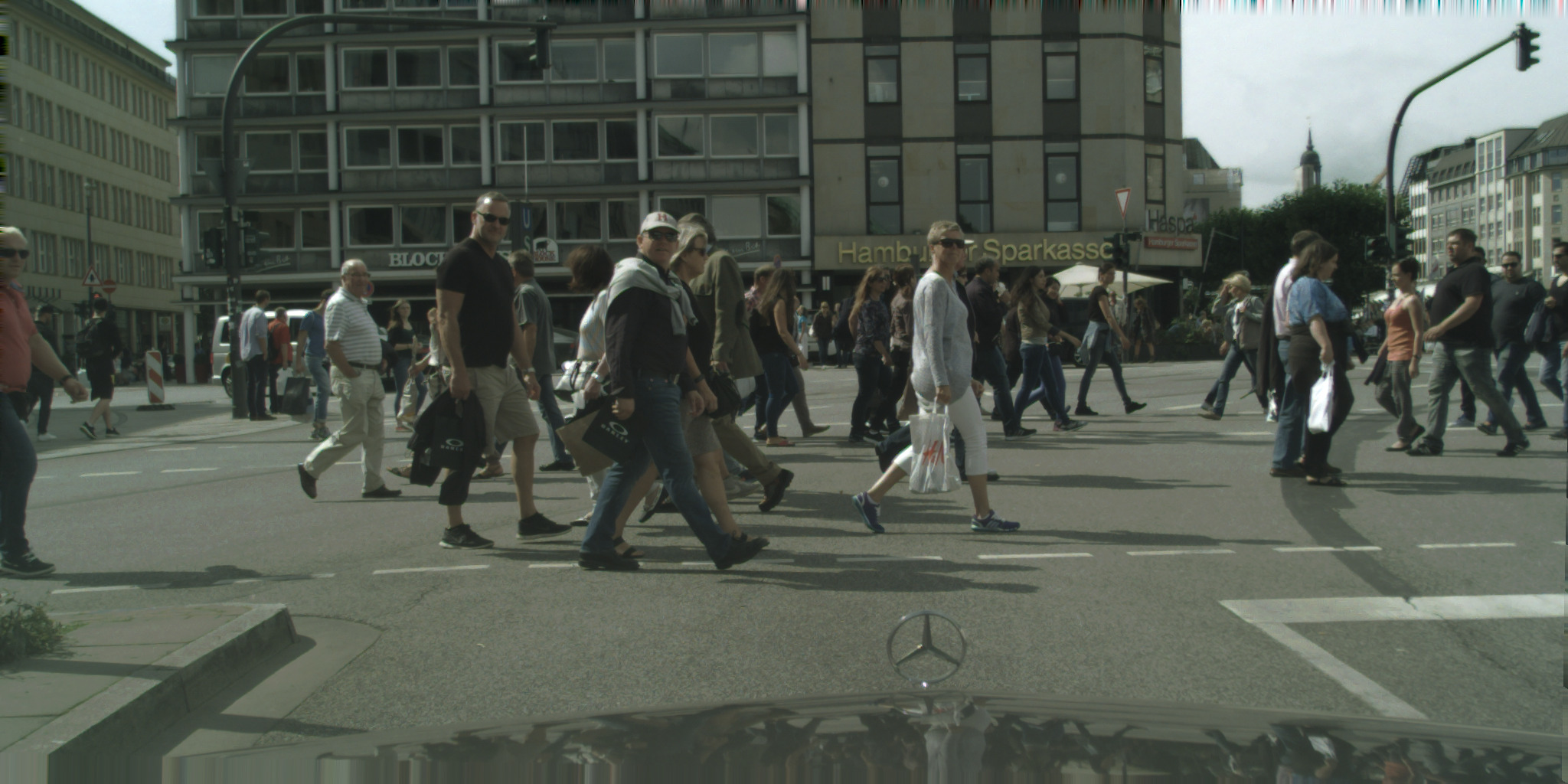}  
		
		\vspace{0.2em}
		
		\includegraphics[width=\linewidth]{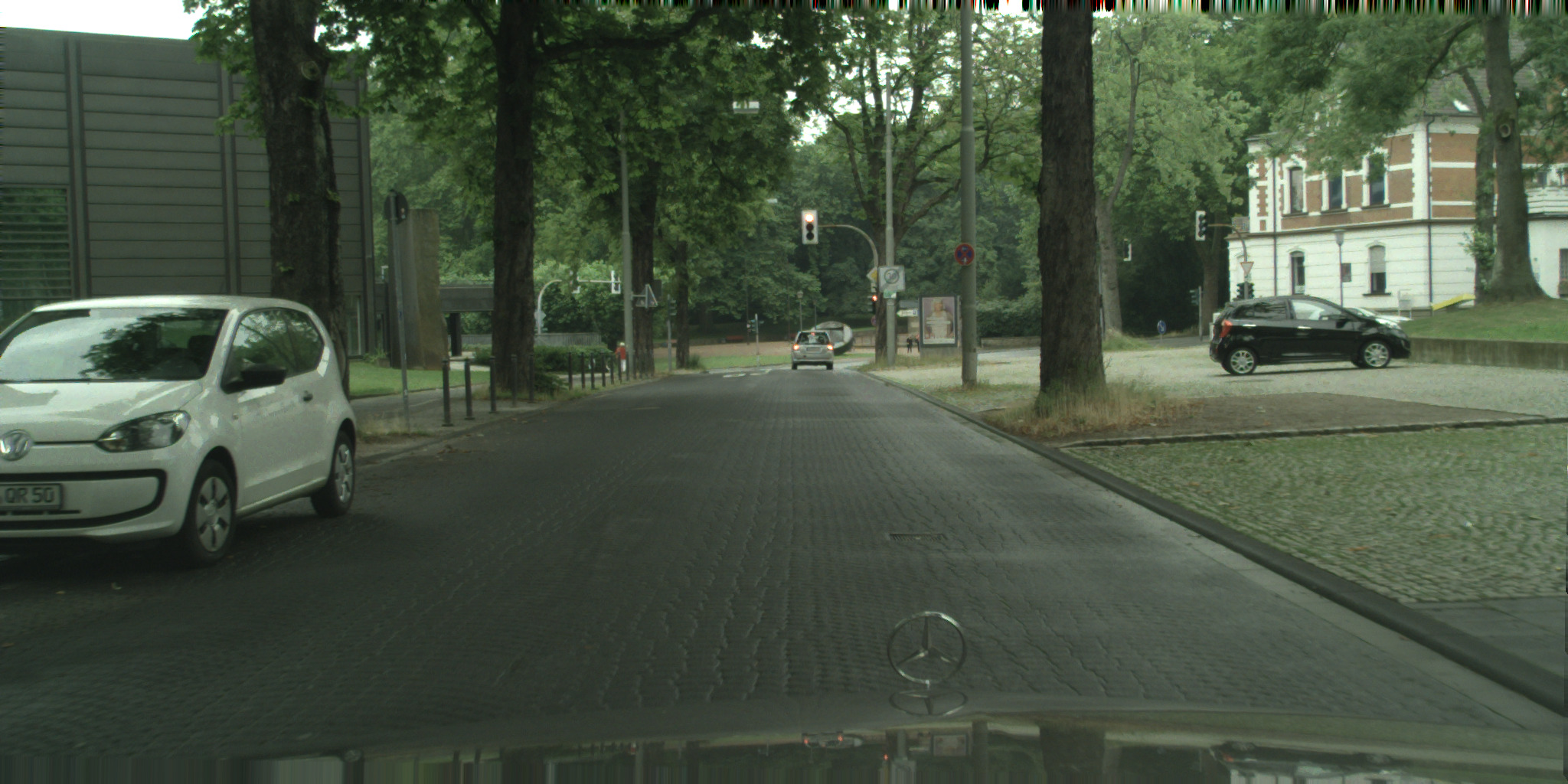}
		
		\vspace{0.2em}
		
		\includegraphics[width=\linewidth]{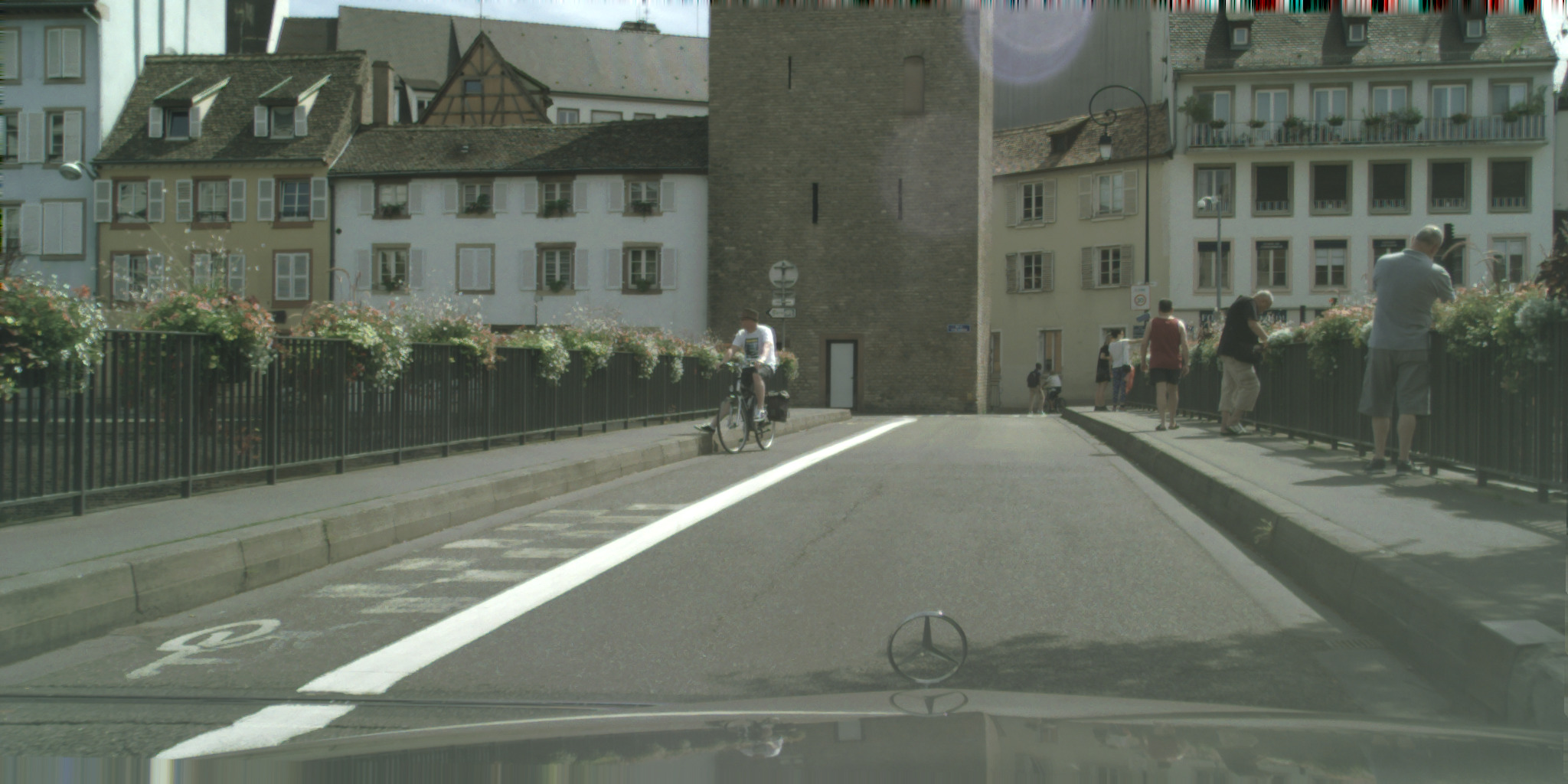}
		
		\vspace{0.2em}
		
		\includegraphics[width=\linewidth]{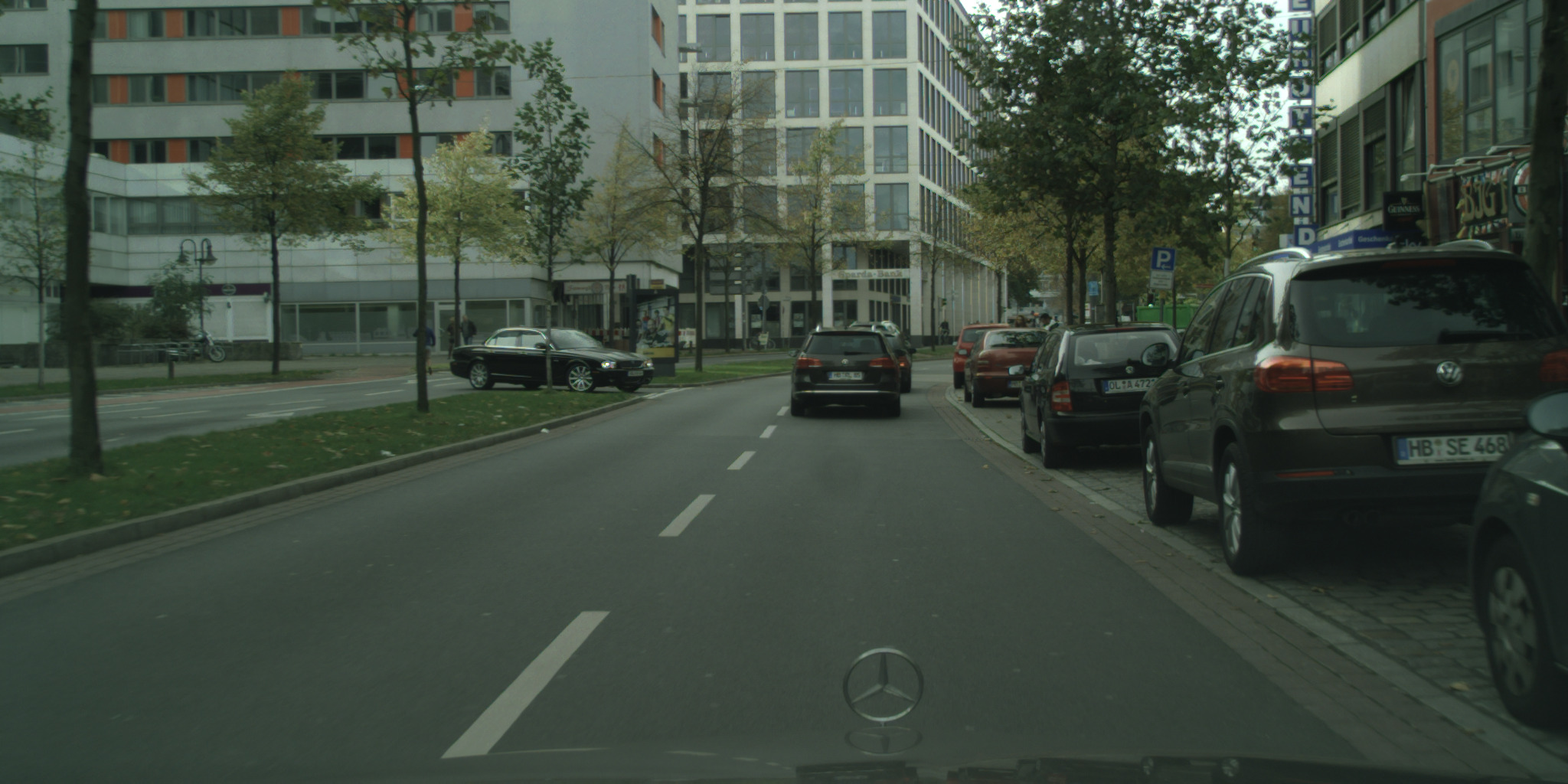}
		\caption{Example images from Cityscapes.}
		\label{fig:cityscapes_input_1}
	\end{subfigure}   
	\hspace{1em}
	\hspace{1em}
	\begin{subfigure}{0.62\linewidth}
		\begin{subfigure}{0.48\linewidth}  
			\centering  
			\includegraphics[width=\linewidth]{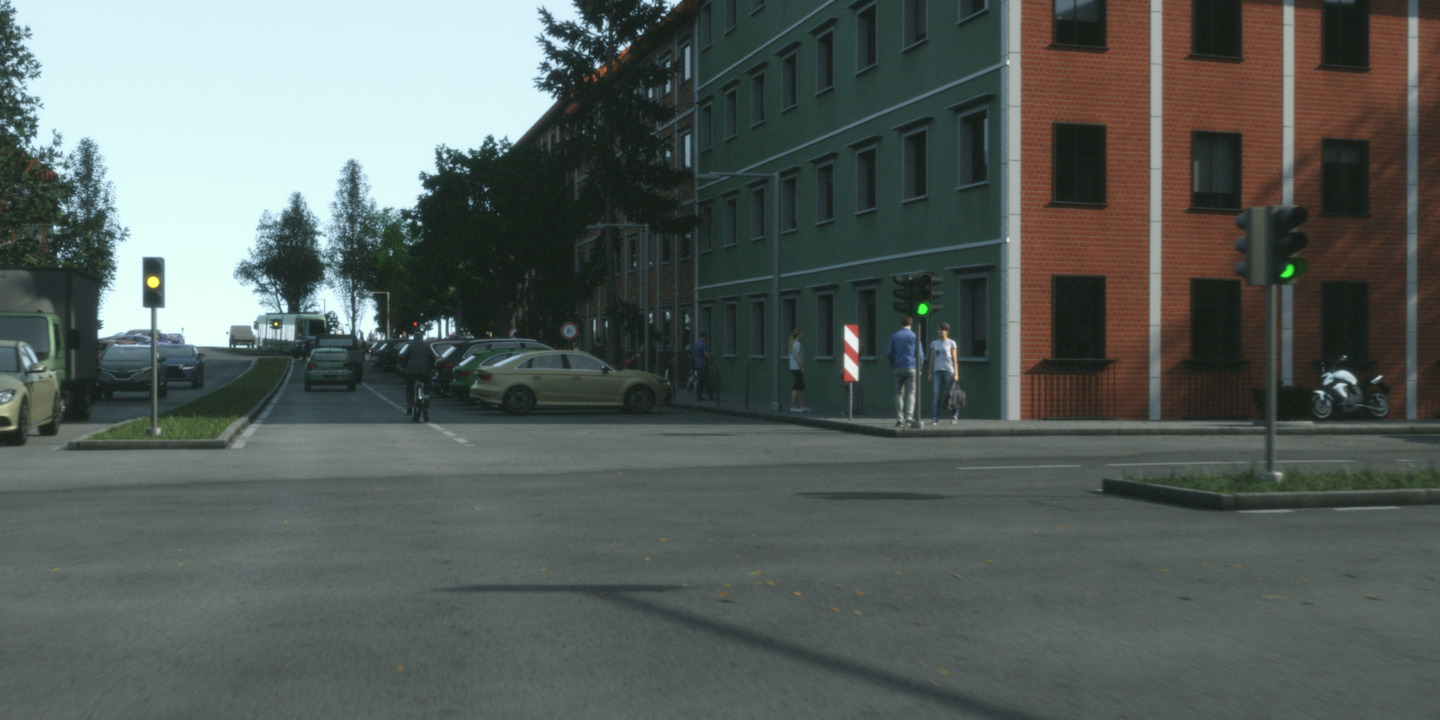}
			
			\vspace{0.2em}
			
			\includegraphics[width=\linewidth]{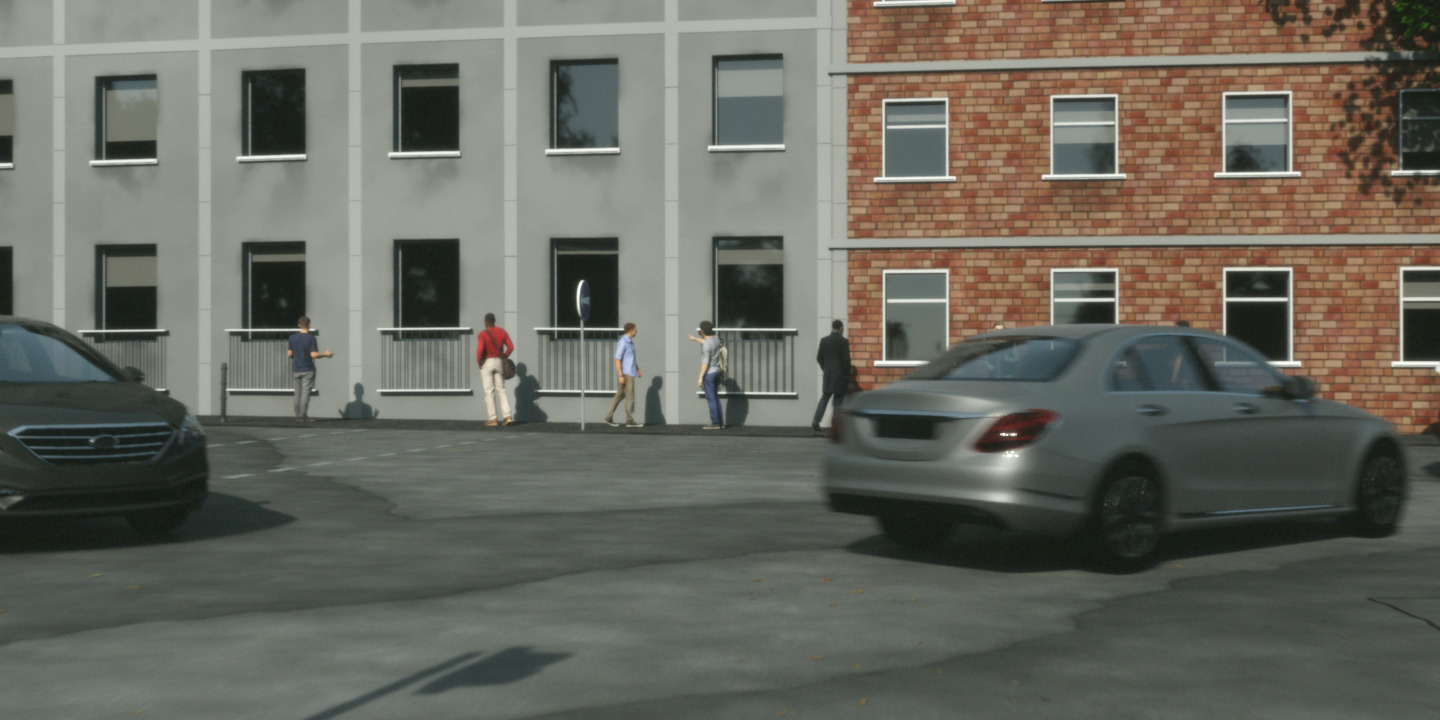}  
			
			\vspace{0.2em}
			
			\includegraphics[width=\linewidth]{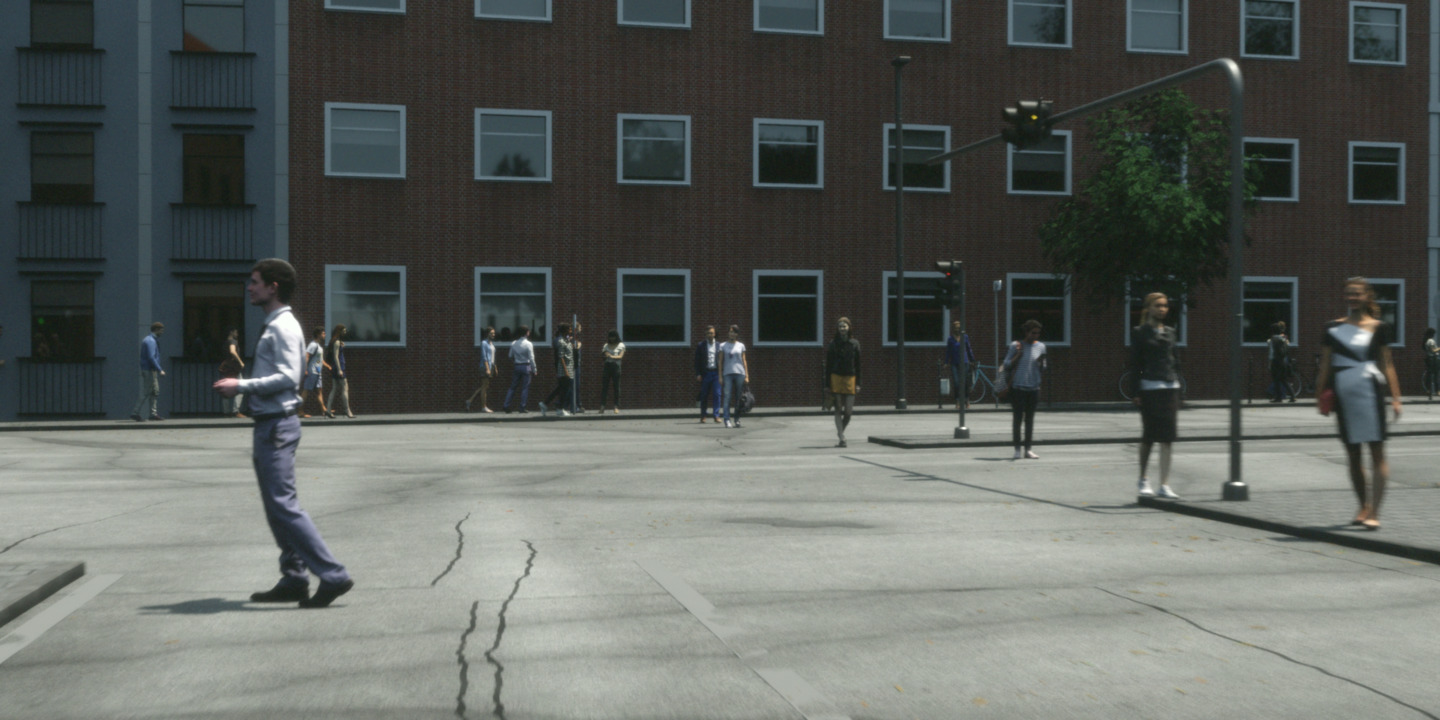}  
			
			\vspace{0.2em}
			
			\includegraphics[width=\linewidth]{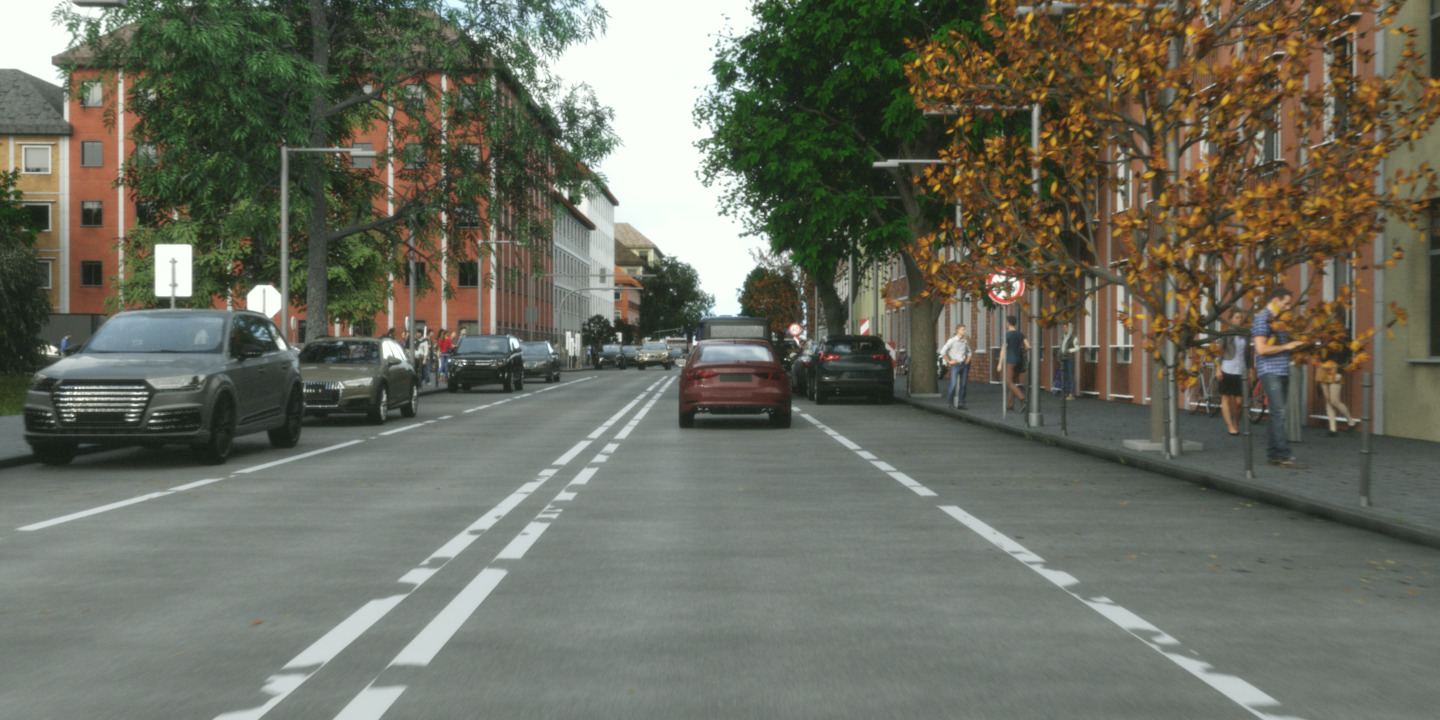}
			
			\vspace{0.2em}
			
			\includegraphics[width=\linewidth]{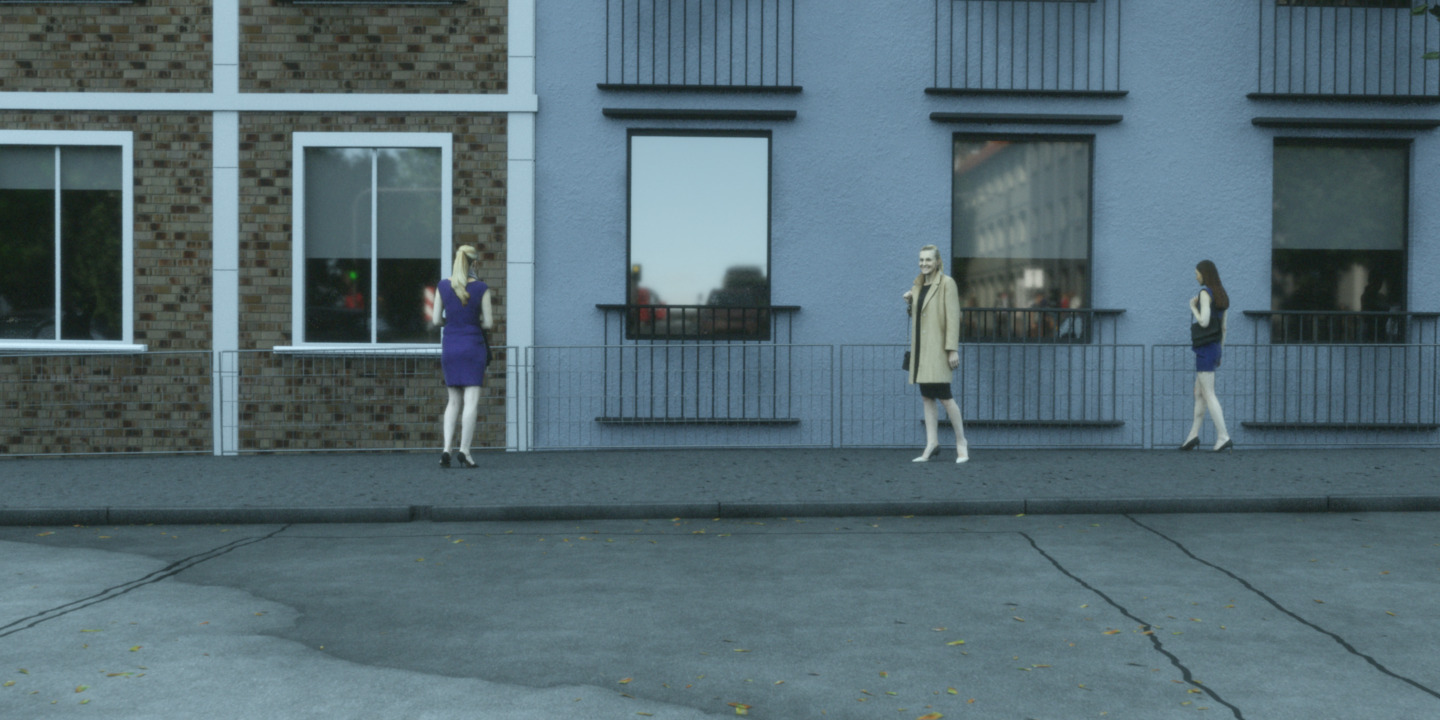}
			
			\vspace{0.2em}
			
			\includegraphics[width=\linewidth]{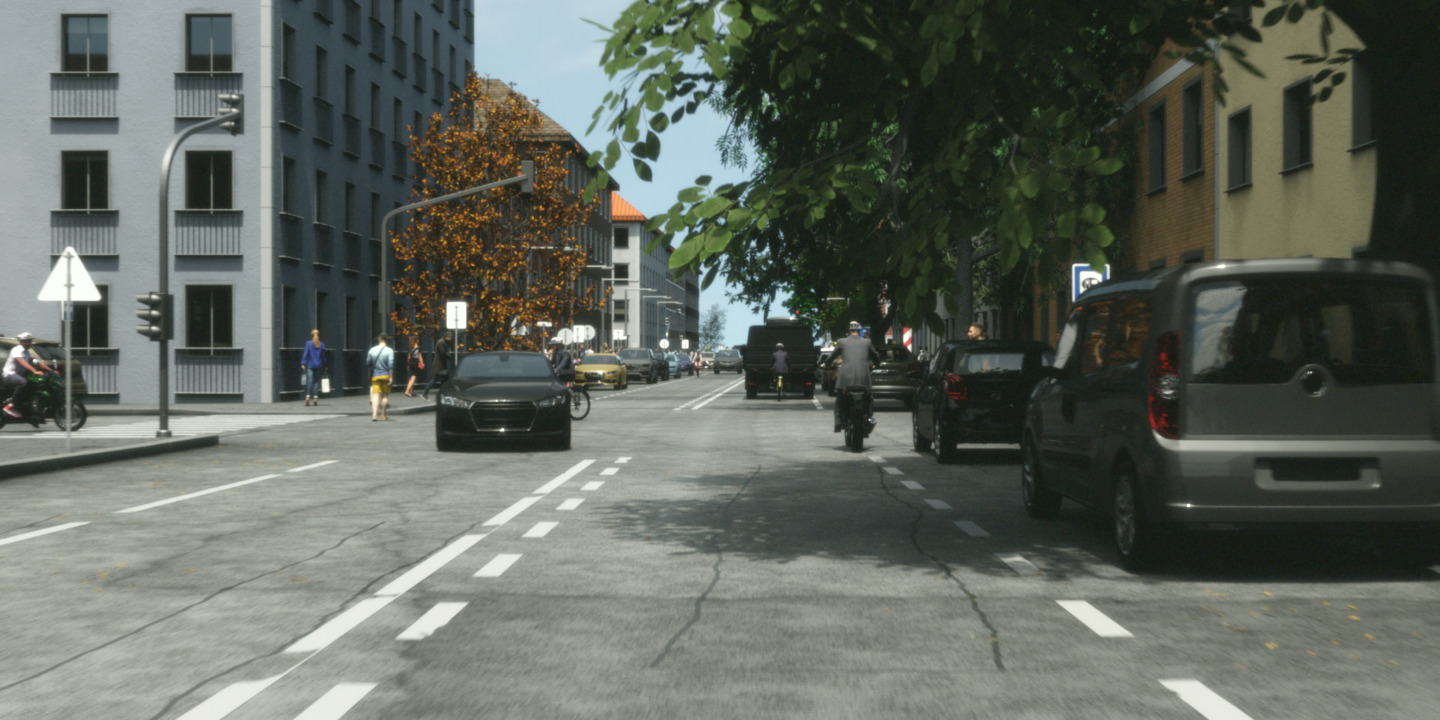}
		\end{subfigure}	
		\hspace{1em}
		\hspace{1em}
		\begin{subfigure}{0.48\linewidth}  
			\centering  
			\includegraphics[width=\linewidth]{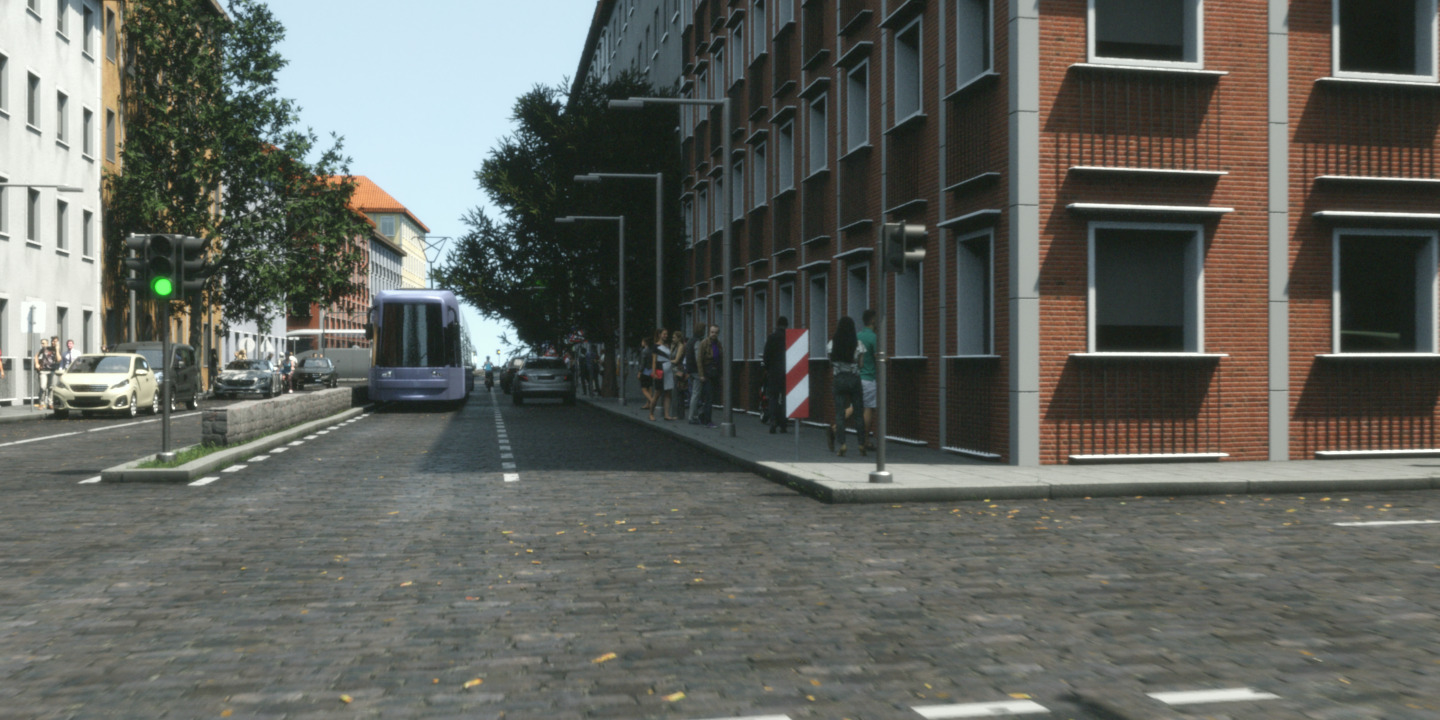}
			
			\vspace{0.2em}
			
			\includegraphics[width=\linewidth]{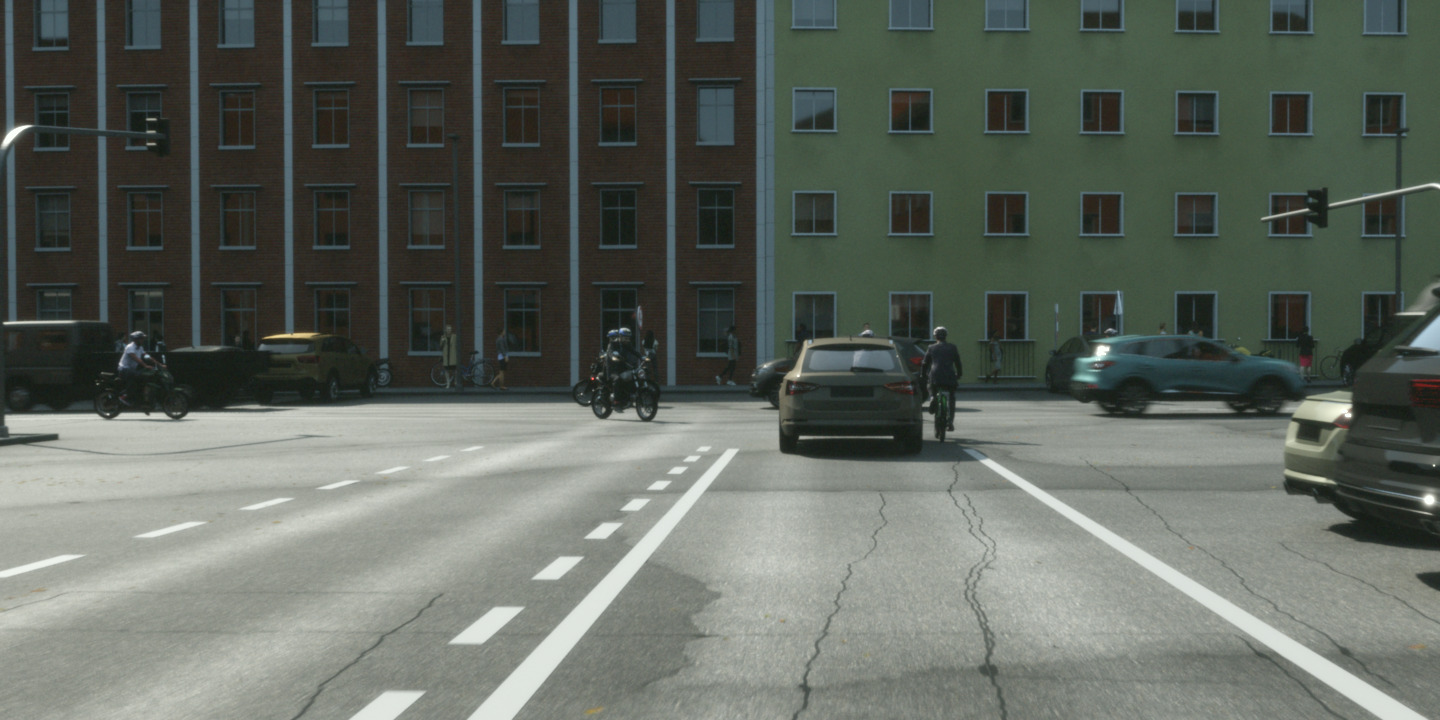}  
			
			\vspace{0.2em}
			
			\includegraphics[width=\linewidth]{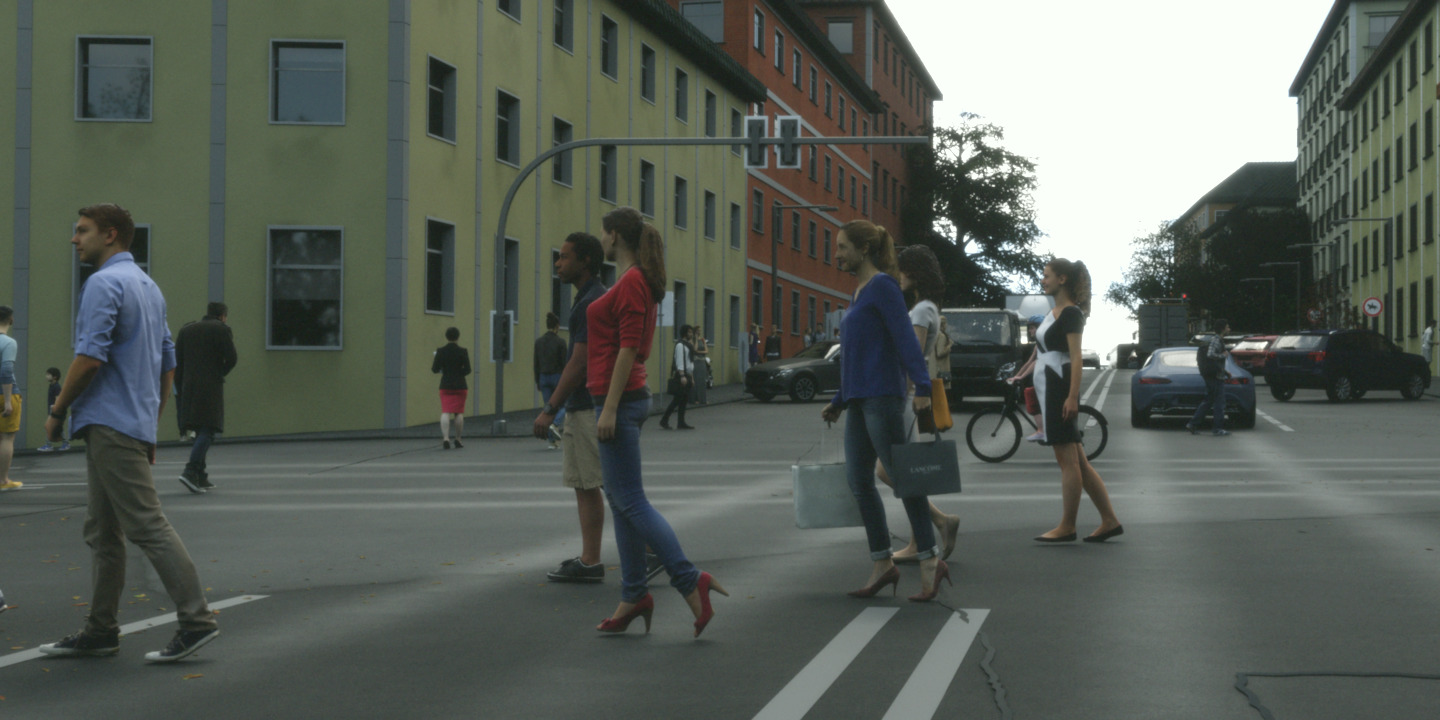}  
			
			\vspace{0.2em}
			
			\includegraphics[width=\linewidth]{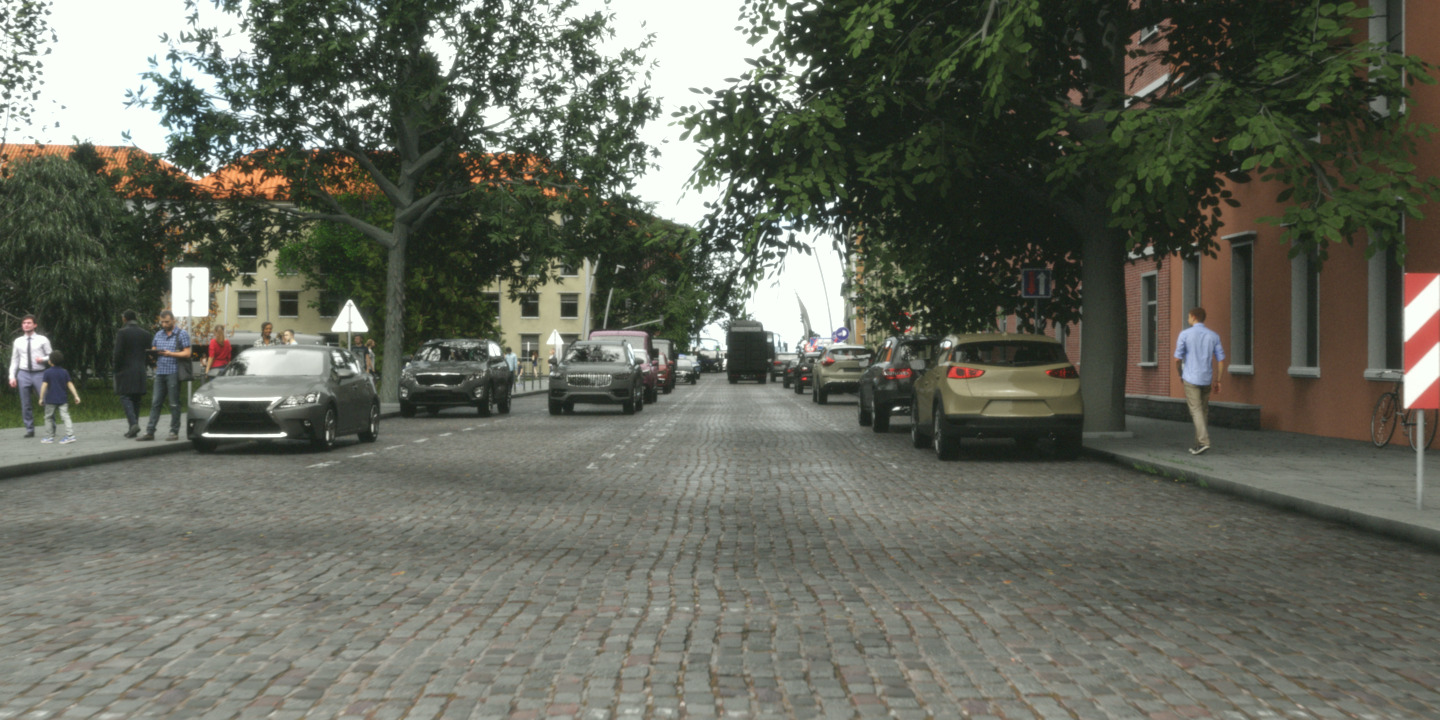}  
			
			\vspace{0.2em}
			
			\includegraphics[width=\linewidth]{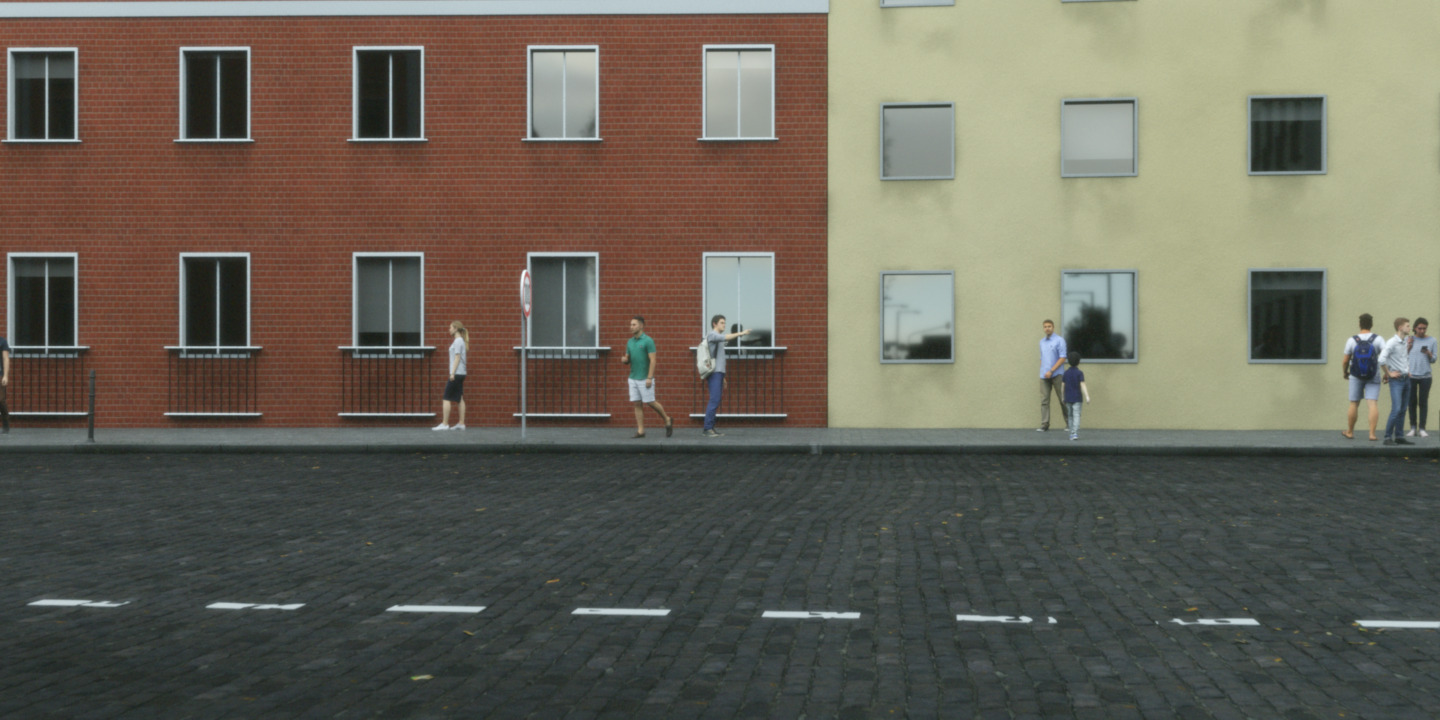}  
			
			\vspace{0.2em}
			
			\includegraphics[width=\linewidth]{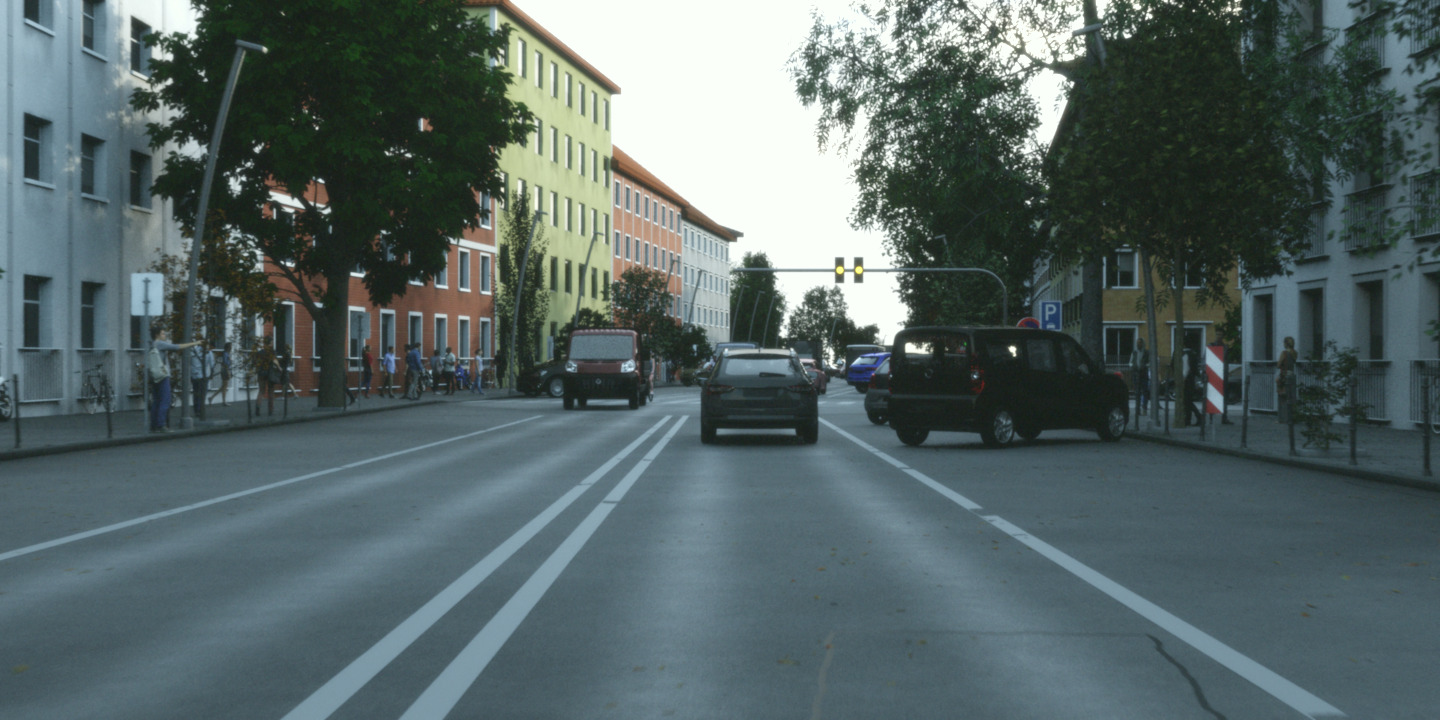}
		\end{subfigure}
		\caption{Semantically similar Synscapes images using Spatial Pyramid Features (SP-feat).}
		\label{fig:spfeat_synscapes_1}
	\end{subfigure}
	\caption{Examples of pairs of images from Cityscapes and Synscapes for training UNIT-GAN. For each example image from Cityscapes (a), we show two example paired-image from Synscapes selected by using our \textit{SP-Feat based selection strategy (b)}. Layout such as position of building, vegetation, pedestrian, vehicles, show the semantic resemblance between real world and synthetic pairs retrieved using \textit{SP-Feat}. Best viewed in color and by zooming in.}
	\label{fig:qualitative_result_5}
\end{figure*}

In order to incorporate semantic information into our model, instead of training UNIT with random pair of images ${x_s\in X_s, x_r\in X_r}$ of the two domains synthetic and real, we used SP-Feat to batch from a set of semantically similar pairs ${x_{s}^{sim}\in X_{s}, x_{r}^{sim}\in X_{r}}$ such that, $x_{s}^{sim} \sim x_{r}^{sim}$ are sets of semantically analogous images from the respective domain, obtained using the K-Nearest Neighbor algorithm on SP-Feat. Note that here only the batch loader is modified and no network architectural changes are being made. To leverage the available real world data information to the maximum, we choose real images as query and find their nearest neighbor in the synthetic domain using SP-Feat. Furthermore, to make the training robust, every synthetic image is associated with a weight inversely proportional to the number of real images it is a neighbor of. Initially every synthetic image is equally probable to get selected. Consequently, if one particular image appears in the neighborhood of multiple query images, we down weigh such an image and replace it with the next Nearest Neighbor image (of higher weight) for subsequent query real image. Using such an approach we choose $k$-Nearest Neighbor for every query image, from which we sample pairs of real image and any of its Nearest Neighbor synthetic image using the smart batch loader. As a trade-off between maximizing the coverage, and not to lose semantic consistency between the query and its $k$-th neighbor, we choose $K=15$ from our analysis. Also keeping in mind the difference in the number of real and synthetic images ($3K$ versus $25K$), we could cover $\approx 70\%$ of synthetic images. In figure \ref{fig:qualitative_result_5} we show qualitative results of such chosen pairs. As shown in figure \ref{fig:spfeat_output}, this approach gets rid of the semantic aberrations that were visible previously. 

In the experiments section we show how the above approach reduces the domain discrepancy between real and synthetic data. It is reflected as an improvement in the downstream task of semantic segmentation for actual-world traffic scenarios, besides better visual quality of the translated images. Further, one of the obvious gains of using synthetic data in network training is to boost the under-represented classes, which otherwise occur rarely in real data. We also show how the above mentioned method plays a vital role in improving the accuracy of semantically classifying such classes using the same segmentation network.

\section{Experimentation, Results and Discussion}

In this section we carry out both quantitative and qualitative evaluations to show the benefits of the approaches to inducing photorealism to synthetic data mentioned in the previous section, in terms of both perceived image quality and domain adaptation. Within the class of most popular scene understanding tasks, semantic segmentation provides fine-grained inference by making dense predictions for every pixel, so that each pixel is labeled with the class it is associated with. We use Cityscapes as representative of real world dataset, and two datasets, Synscapes and GTA-V as synthetic data in our experiments. Since both the synthetic datasets contain around $25,000$ annotated  images, compared to just $3,000$ fine-label annotated Cityscapes' images, we randomly sample equal number of $3,000$ images from both of it to maintain parity between the different datasets. Also, to remove any training induced bias in the evaluation, we keep all the hyperparameters fixed for all semantic segmentation experiments. Like batch-size, number of epochs, learning-rate scheduler and optimizer. We use the open-sourced ERF-Net model \cite{erfnet} for this task, because of its efficiency and ease of training. The images are resized to $256 \times 512$ dimension, prior to feeding into the network while training.

We would like to emphasize that the main motive of our work is to demonstrate how to improve domain translation of images between real and synthetic, and not obtain segmentation SOTA accuracy. Translation sometimes may induce artifacts in the image which leads to improper label assignment. E.g., \textit{sky} may get inappropriately labelled as \textit{tree} because of such induced noises (Fig. \ref{fig:unit_gan_output}). We show how our SP-Feat based pairing approach alleviates this. Semantic segmentation is simply chosen as a downstream task to validate this improvement. Any improvement on the baseline segmentation performance obtained using hyper-parameter tuning or a better network, will translate to equivalent gains using our smart batch loader approach, since our proposed method is agnostic of the subsequent segmentation model.

\begin{table}
	\small
	\begin{center}
		\begin{tabular}{|c|c|}
			\hline
			Training set & Mean IoU (\%)\\
			\hline\hline
			GTA-V & 16.5 \\
			Photorealistic GTA-V & \textbf{27.13} \\
			\hline
			Synscapes & 41 \\
			Photorealistic Synscapes & \textbf{43} \\
			\hline
			Cityscapes \textit{(baseline)} & 70.06 \\
			Cityscapes + Photorealistic GTA-V & \textbf{71.2} \\
			Cityscapes + Photorealistic Synscapes & \textbf{71.75} \\
			\hline
		\end{tabular}
	\end{center}
	\caption{Semantic Segmentation results using ERF-Net on Cityscapes Validation set.}
	\label{table:sem_seg_baseline}
\end{table}

In table \ref{table:sem_seg_baseline} we show the advantage of domain gap reduction between Cityscapes and the synthetic data by training the network only on synthetic data, but validating on the validation set of real-world Cityscapes data. From the results it is evident that the gains obtained by inducing photo-realism is more prominent in case of GTA-V, compared to Synscapes. The reason can be attributed to the high quality of the synthetic data already available in case of Synscapes, when compared to GTA-V. But after combining with Cityscapes training data with the induced synthetic data, we obtain at-par accuracy using either of the synthetic dataset. This demonstrates the trade-off between time and resource spent in generating higher quality synthetic data, and using our approach to benefit from lower quality synthetic data as well.

To better understand the gains obtained, we append the Cityscapes training images with equal number of photorealistic Synscapes images, and compare against the baseline model trained on only Cityscapes. We compare both original UNIT-GAN  and our proposed \textit{smart batch-loader} approach for translating Synscapes. Table \ref{table:classwise_accuracy} shows that overall our method performs better than the baseline and basic UNIT-GAN  approach.

\begin{table*}
	\begin{center}
		\resizebox{\textwidth}{!}
		{%
			
			\begin{tabular}{|c|c|c|c|}
				\hline
				\multirow{3}*{Class} & \multicolumn{3}{|c|}{Cityscapes Validation IoU\% when Trained on} \\
				\cline{2-4}
				& \multirow{2}*{Cityscapes \textit{(baseline)}} & \multirow{2}*{Cityscapes + UNIT-GAN Synscapes} & Cityscapes + Smart Batch-loader \\
				& & & UNIT-GAN Synscapes (ours) \\
				\hline \hline
				Road & 97.26 & 97.5 & \textbf{97.66} \\
				\hline
				Sidewalk & 80.03 & 80.95 & \textbf{82.26} \\
				\hline
				Building & 90.42 & 90.69 & \textbf{90.96} \\
				\hline
				Wall & 48.87 & 48.76 & \textbf{53.76} \\
				\hline
				Fence & 52.18 & 53.74 & \textbf{55.98} \\
				\hline
				Pole & 57.55 & 59.85 & \textbf{60.95} \\
				\hline
				Traffic-light & 60.27 & \textbf{62.88} & 62.62 \\
				\hline
				Traffic-sign & 70.16 & 70.56 & \textbf{71.79} \\
				\hline
				Vegetation & 91.11 & 91.1 & \textbf{91.29} \\
				\hline
				Terrain & \textbf{63.08} & 60.71 & 62.80 \\
				\hline
				Sky & \textbf{93.66} & 93.26 & 93.14 \\
				\hline
				Person & 74.97 & 74.27 & \textbf{76.60} \\
				\hline
				Rider & 52 & 55.45 & \textbf{56.95} \\
				\hline
				Car & 91.42 & 92.36 & \textbf{93.2} \\
				\hline
				Truck & 59.69 & 69.89 & \textbf{72.28} \\
				\hline
				Bus & 70.63 & 76.26 & \textbf{79.61} \\
				\hline
				Train & 64.11 & \textbf{69.62} & 65.35 \\
				\hline
				Motorcycle & 45.33 & 47.02 & \textbf{58.32} \\
				\hline
				Bicycle & 68.44 & 68.27 & \textbf{70.55} \\
				\hline \hline
				\textbf{Mean IoU} & 70.06 & 71.75 & \textbf{73.48} \\
				\hline
			\end{tabular}%
		}
	\end{center}
	\caption{Semantic Segmentation performance using Photorealistic Synscapes and Cityscapes training data.}
	\label{table:classwise_accuracy}
\end{table*}

\begin{figure}
	\centering
	\includegraphics[width=0.5\linewidth]{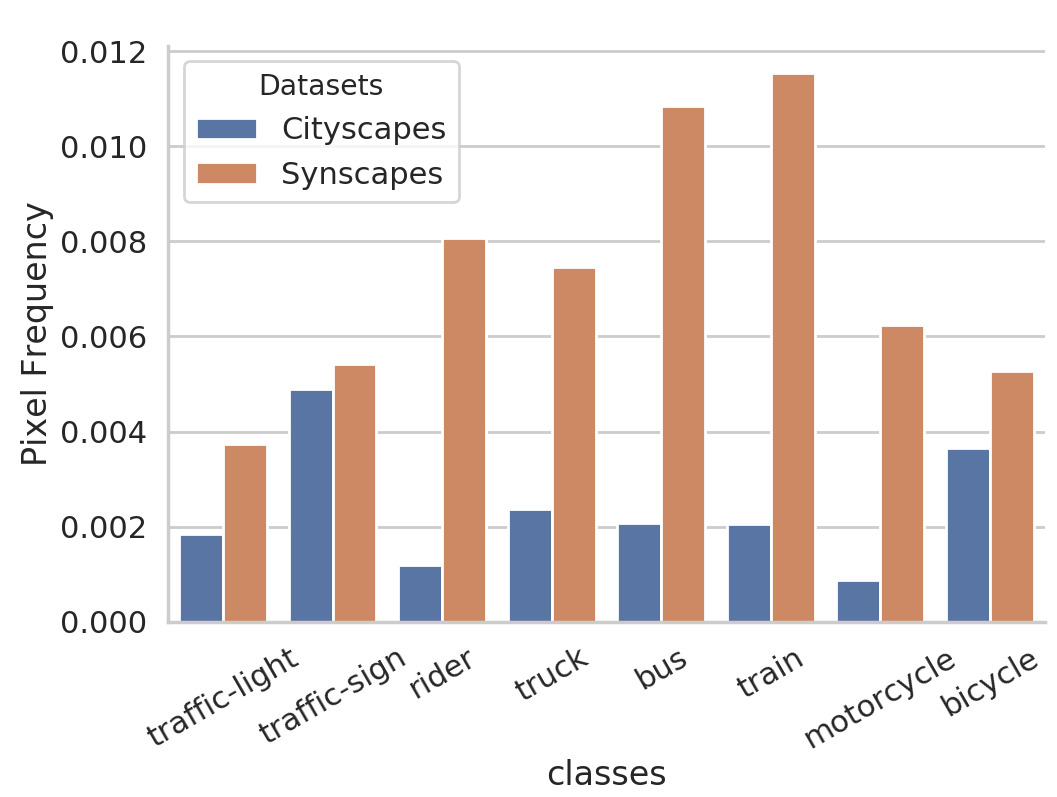}
	\caption{Relative pixel count comparison for under-represented classes in Cityscapes training set and equal number of sampled Synscapes images.}
	\label{fig:pixel_count}
\end{figure}

\begin{table}
	\small
	\begin{center}
		\begin{tabular}{|c|c|c|c|}
			\hline
			\multirow{3}*{Class} & \multicolumn{3}{|c|}{Validation IoU\% when Trained on} \\
			\cline{2-4}
			& Cityscapes & Cityscapes \& & Cityscapes \& PR \\
			& & Synscapes & Synscapes (ours) \\
			\hline \hline
			Traffic light & 60.27 & 61.59 & \textbf{62.62} \\
			\hline
			Traffic-sign & 70.16 & 71.21 & \textbf{71.79} \\
			\hline
			Rider & 52 & 56.38 & \textbf{56.95} \\
			\hline
			Truck & 59.69 & 72.04 & \textbf{72.28} \\
			\hline
			Bus & 70.63 & 77.5 & \textbf{79.61} \\
			\hline
			Train & 64.11 & \textbf{67.46} & 65.35 \\
			\hline
			Motorcycle & 45.33 & 44.67 & \textbf{58.32} \\
			\hline
			Bicycle & 68.44 & 69.94 & \textbf{70.55} \\
			\hline
		\end{tabular}
	\end{center}
	\caption{Performance on under-represented classes using Photorealistic (PR) Synscapes and Cityscapes training data.}
	\label{table:low_class_accuracy}
\end{table}

In the context of semantic segmentation, one of the advantages of using synthetic data is to increase the instances containing the under-represented classes of actual data. In figure \ref{fig:pixel_count} we compare the relative pixel-counts for these minority classes. It is evident that in the sampled Synscapes images the relative pixel density for those classes are more than that in Cityscapes. Table \ref{table:low_class_accuracy} shows that boosting the density of the under-represented classes improves performance of the model on them as well. The smart pairing of images, as explained above, further improves the performance.

The gains obtained by making the synthetic data more photorealistic can be seen as reduction in domain gap between the two domains. Although evaluating performance on semantic segmentation clearly shows the advantage of such an approach, we want to have more detailed understanding about the difference in the distribution. In order to do so we use the \textit{Fréchet Inception Distance (FID)} to measure the difference between feature distributions.

\textit{\textbf{Fréchet Inception Distance (FID):}} It is a popular metric in the domain of image generation, for measuring the difference in feature distribution of generated image samples and real samples. It does so by using the statistics of the activation of ImageNet \cite{imagenet} pre-trained Inception-V3 network \cite{inceptionv3}, for each domain. Lower FID between two data distributions signify more similarity in their feature space.

\begin{table}
	\small
	\begin{center}
		\begin{tabular}{|c|c|c|}
			\hline
			Reference set & Cityscapes & Synscapes \\
			\hline \hline
			Baseline & & \\
			\hline \hline
			5k Cityscapes & 3.32 & 43.44 \\
			5k Synscapes & 43.44 & 3.93 \\
			\hline \hline
			Photorealistic Synscapes using & & \\
			\hline \hline
			Auto-encoder & 39.91 & 6.28 \\
			Cycle GAN & 28.71 & 12.53 \\
			UNIT-GAN & 26.01 & 13.22 \\
			Smart Batch-Loader (ours) & \textbf{20.62} & \textbf{13.813} \\
			\hline
		\end{tabular}
	\end{center}
	\caption{FID of Photorealistic synthetic data in reference to Real-world and Synthetic data. Lower w.r.t Cityscapes is better and Higher w.r.t Synscapes.}
	\label{table:fid}
\end{table}

While computing FID, we randomly select $5000$ images from each set, and resize them to $256\times512$. We also compare it against randomly selected mutually exclusive set of $5000$ images from both Cityscapes and Synscapes. Lower FID score with respect to Cityscapes denotes better adoption of the synthetic data to the real-world distribution. Semantically consistent pair formation is a pre-processing step, and not tightly coupled to the training of the image translation network. This enables it to be used across architectures. In our paper we choose UNIT-GAN since it is one of the SOTA networks, combining the advantages of many preceding models. To validate this claim, we compare against two other widely used approaches for domain translation, namely auto-encoder based architecture without having the discriminator based adversarial loss, and normal Cycle-GAN \cite{cyclegan} to convert Synscapes images to Cityscapes domain. As can be seen from table \ref{table:fid}, it is indeed the case and hence our usage of the best model, that is UNIT-GAN, to demonstrate the added benefit of our approach. Our proposed method is able to induce photorealism in synthetic data better than the baseline UNIT-GAN framework as is evident from the lowest FID with respect to Cityscapes. The synthetic data is gradually able to get out of the synthetic domain distribution and adapt to the real world distribution. Hence, there is a gradual increase of the FID when compared with the Synscapes images. This also matches with our observation related to the semantic segmentation accuracies reported in table \ref{table:classwise_accuracy}. 

\begin{figure*}[t]
	\centering  
	\begin{subfigure}{\linewidth}  
		\centering  
		\includegraphics[width=0.35\linewidth]{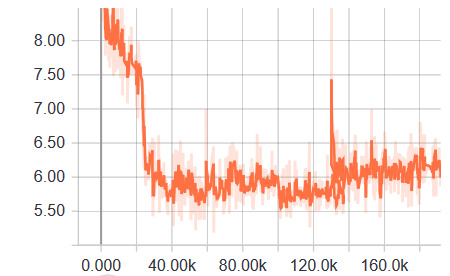}
		\includegraphics[width=0.35\linewidth]{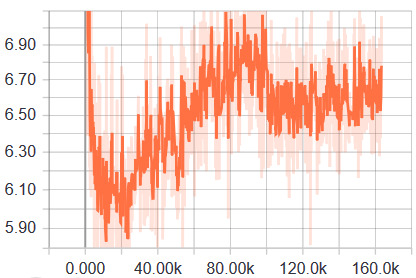}  
		\caption{Generator loss curves of training original (\textit{left}) and \textit{Smart Batch-loader} based (\textit{right}) UNIT-GAN.}
		\label{fig:gan_generator}
	\end{subfigure}   
	
	\begin{subfigure}{\linewidth}  
		\centering  
		\includegraphics[width=0.35\linewidth]{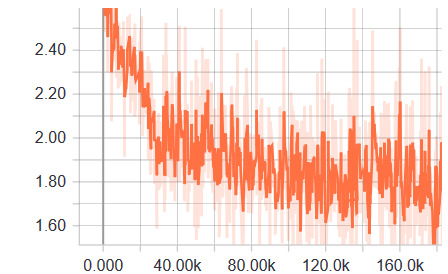}
		\includegraphics[width=0.35\linewidth]{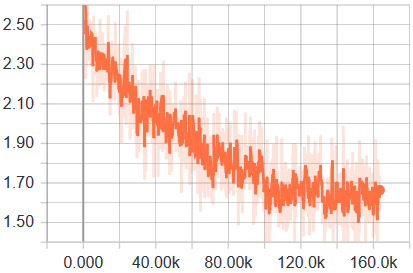}  
		\caption{Discriminator loss curves of training original (\textit{left}) and \textit{Smart Batch-loader} based (\textit{right}) UNIT-GAN.}
		\label{fig:gan_discriminator}
	\end{subfigure}
	\caption{Loss curve of generator and discriminator respectively, for training original (\textit{left}) and \textit{Smart Batch-loader} based (\textit{right}) UNIT-GAN. Smoothing factor of $0.5$ is used for generating all the curves. Best viewed by zooming in.}
	\label{fig:qualitative_result_6}
\end{figure*}

For better insights, we also visualize in figure \ref{fig:qualitative_result_6}, the total loss curves for the generator and the discriminator of the GAN when trained using random pairs of images and using our proposed \textit{smart batch-loader} technique, respectively. In case of the generator, figure \ref{fig:gan_generator} shows that SP-Feat based training operates in a lower range of values $[5.90, 7.10]$, compared to original training ($[5.50, 8.50]$). However, looking at the spiky nature of the curve for SP-Feat based training, it can be deduced that the generator has to work harder to \textit{generate} more realistic looking synthetic images, compared to the former. In training a typical GAN architecture, these can be attributed to better loss being propagated from the discriminator to the generator while working with semantically similar pairs of images. Moreover, the discriminator too demonstrates a much stabler loss curve while being trained with \textit{smart batch-loader}, than the original training curve (\ref{fig:gan_discriminator}). These observations validates our intuition of bettering the GAN training using semantically coherent pairs of images from real and synthetic domain.

\begin{figure*}
	\centering  
	\begin{subfigure}{0.9\linewidth}  
		\centering  
		\includegraphics[width=0.32\linewidth]{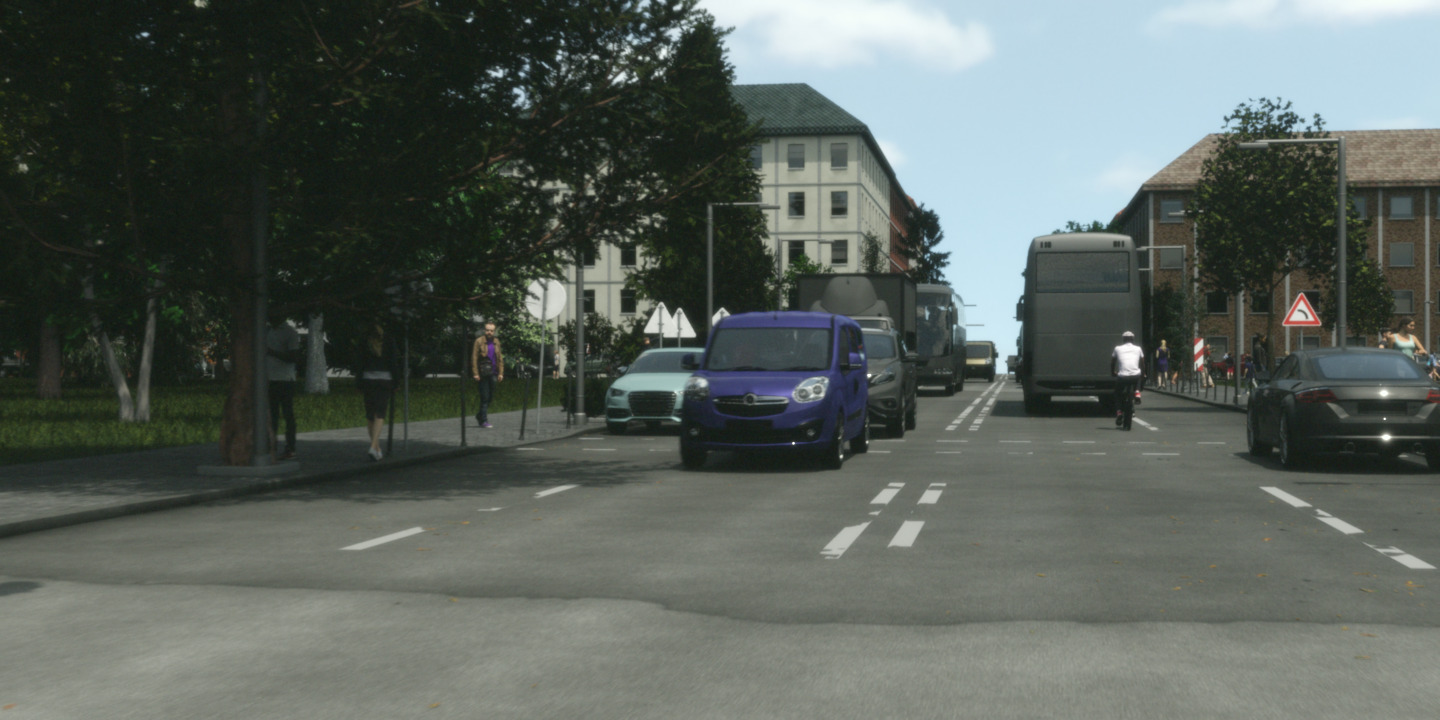}
		\includegraphics[width=0.32\linewidth]{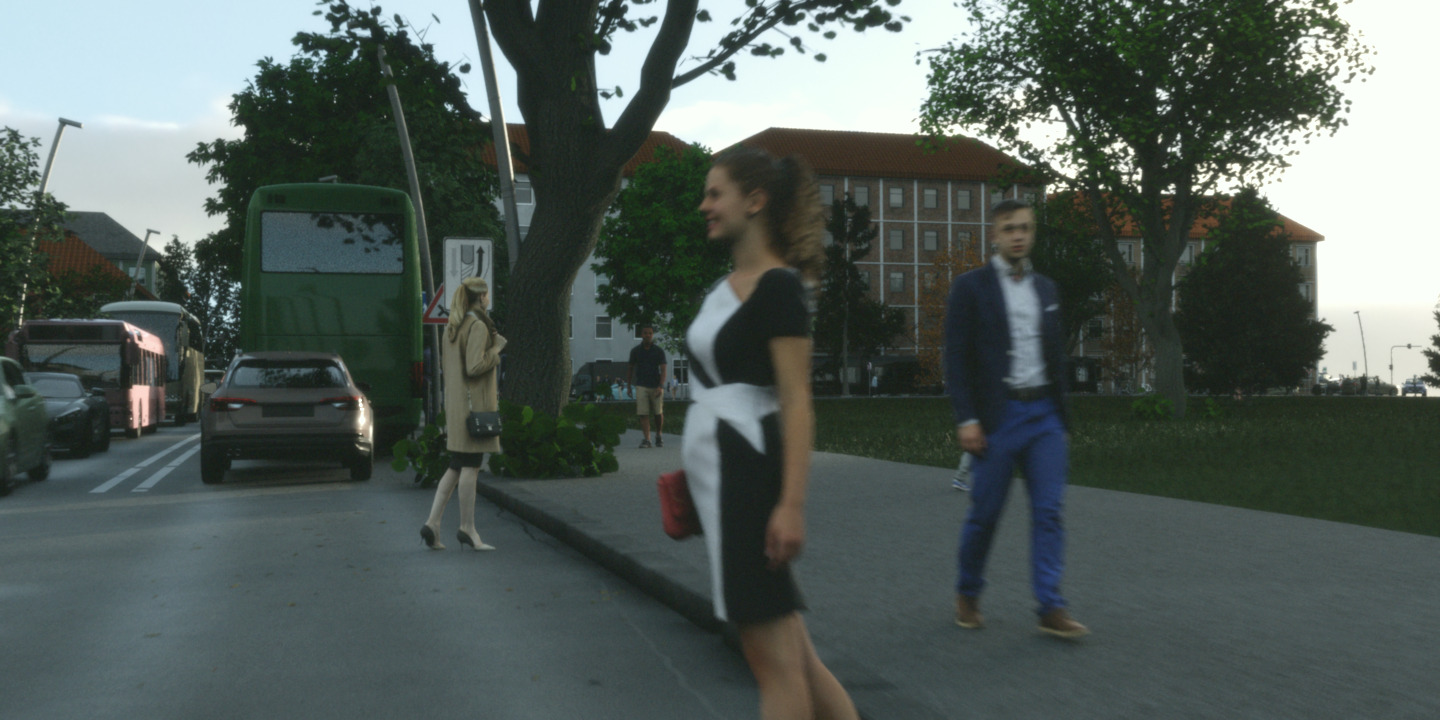}  
		\includegraphics[width=0.32\linewidth]{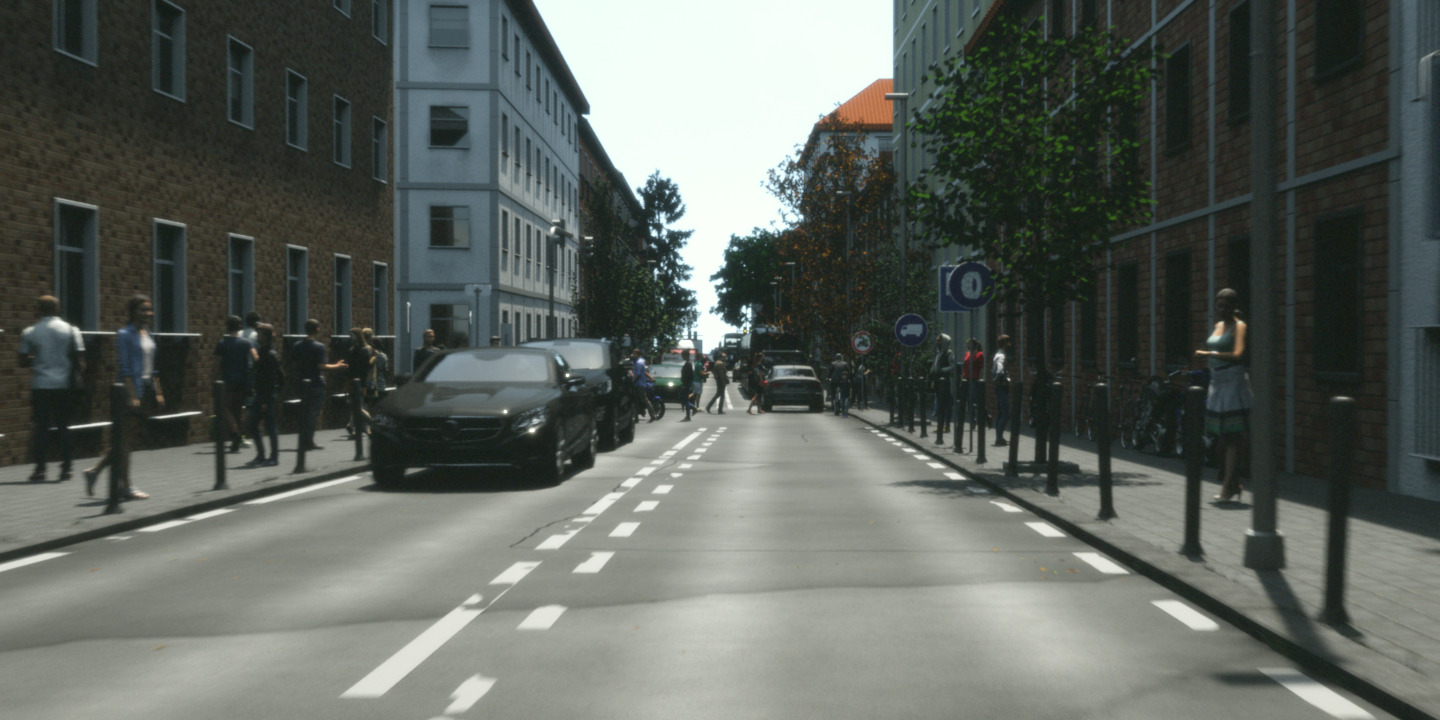}  
		\caption{Example Synthetic images from Synscapes.}
		\label{fig:synscape_input_2}
	\end{subfigure}   
	
	\begin{subfigure}{0.9\linewidth}  
		\centering  
		\includegraphics[width=0.32\linewidth]{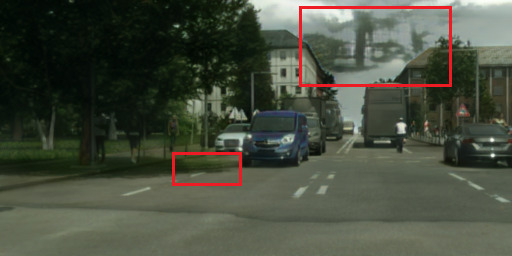}
		\includegraphics[width=0.32\linewidth]{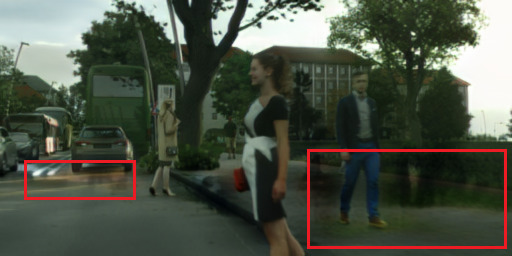}  
		\includegraphics[width=0.32\linewidth]{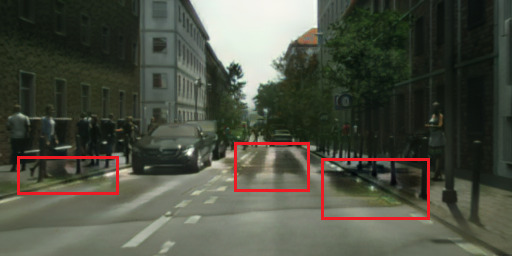}  
		\caption{Results using original UNIT-GAN architecture.}
		\label{fig:unit_gan_output_2}
	\end{subfigure}
	
	\begin{subfigure}{0.9\linewidth}  
		\centering  
		\includegraphics[width=0.32\linewidth]{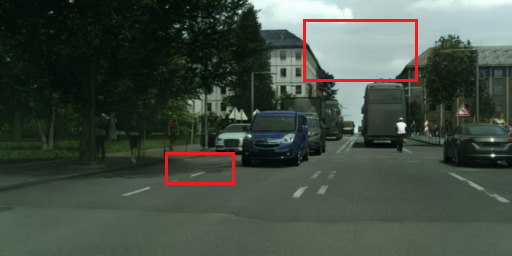}
		\includegraphics[width=0.32\linewidth]{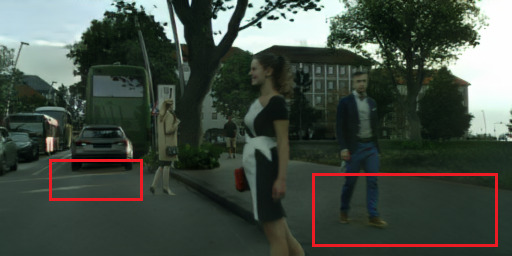}  
		\includegraphics[width=0.32\linewidth]{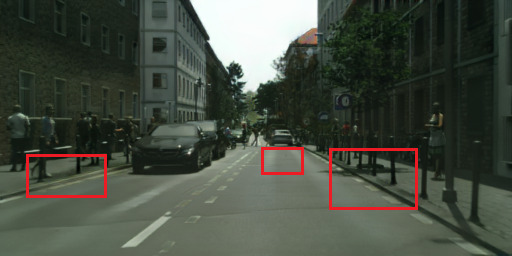}  
		\caption{Results using SP-Feat based Batch selection strategy.}
		\label{fig:spfeat_output_2}
	\end{subfigure}
	
	\caption{Result using Smart Batch-Loader for inducing Photorealism in Synscapes, from Cityscapes. \textit{Red boxes} highlight some regions showing our proposed approach gets rid of noises induced using basic UNIT-GAN network, while inducing photorealism. Best viewed in color.}
	\label{fig:qualitative_result_2}
\end{figure*}

In figure \ref{fig:qualitative_result_2} we show more examples of the improved visual quality of the transformed images obtained using our proposed batch selection strategy. More results are provided in the supplementary material. Analyzing the correlation between visual quality and feature adaptability of photorealistic synthetic images to real data, we empirically observe that the two aspects are correlated.

\begin{figure*}
	\centering  
	\begin{subfigure}{0.9\linewidth}  
		\centering  
		\includegraphics[width=0.32\linewidth]{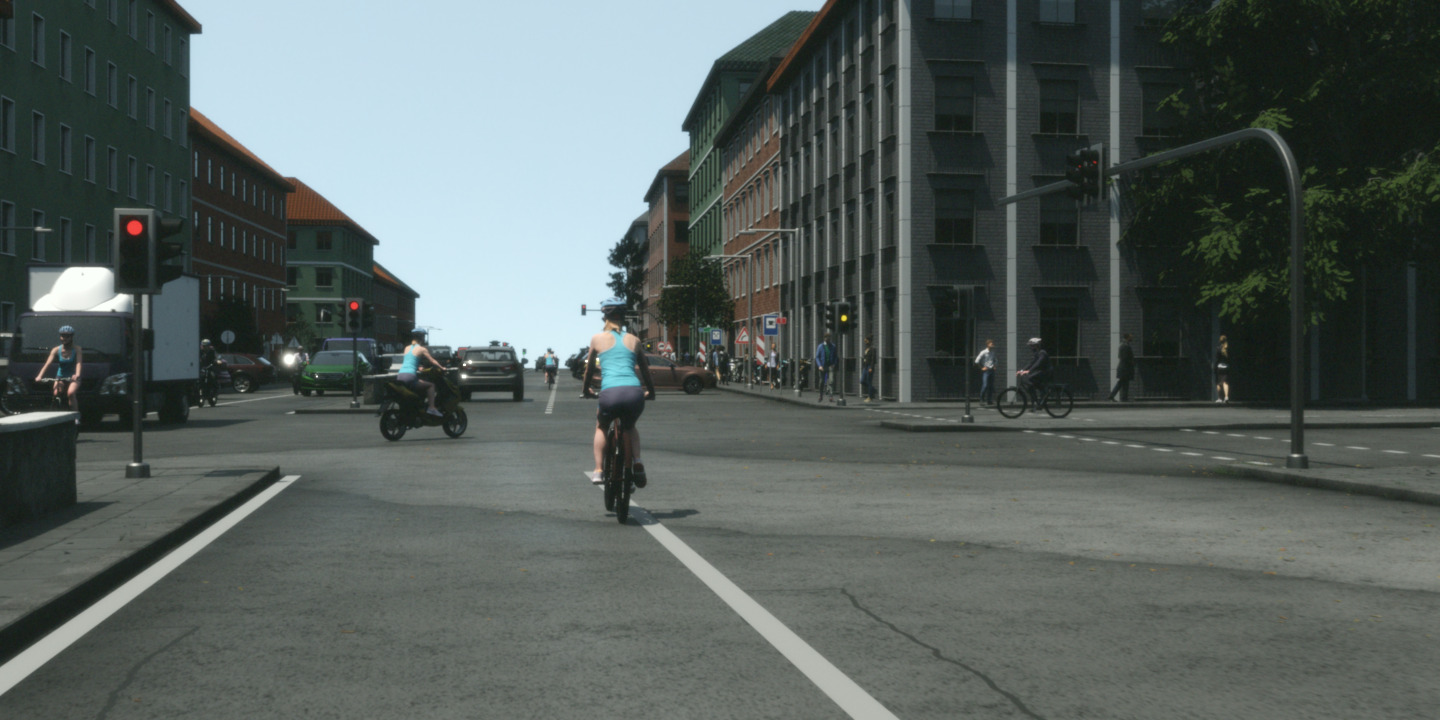}
		\includegraphics[width=0.32\linewidth]{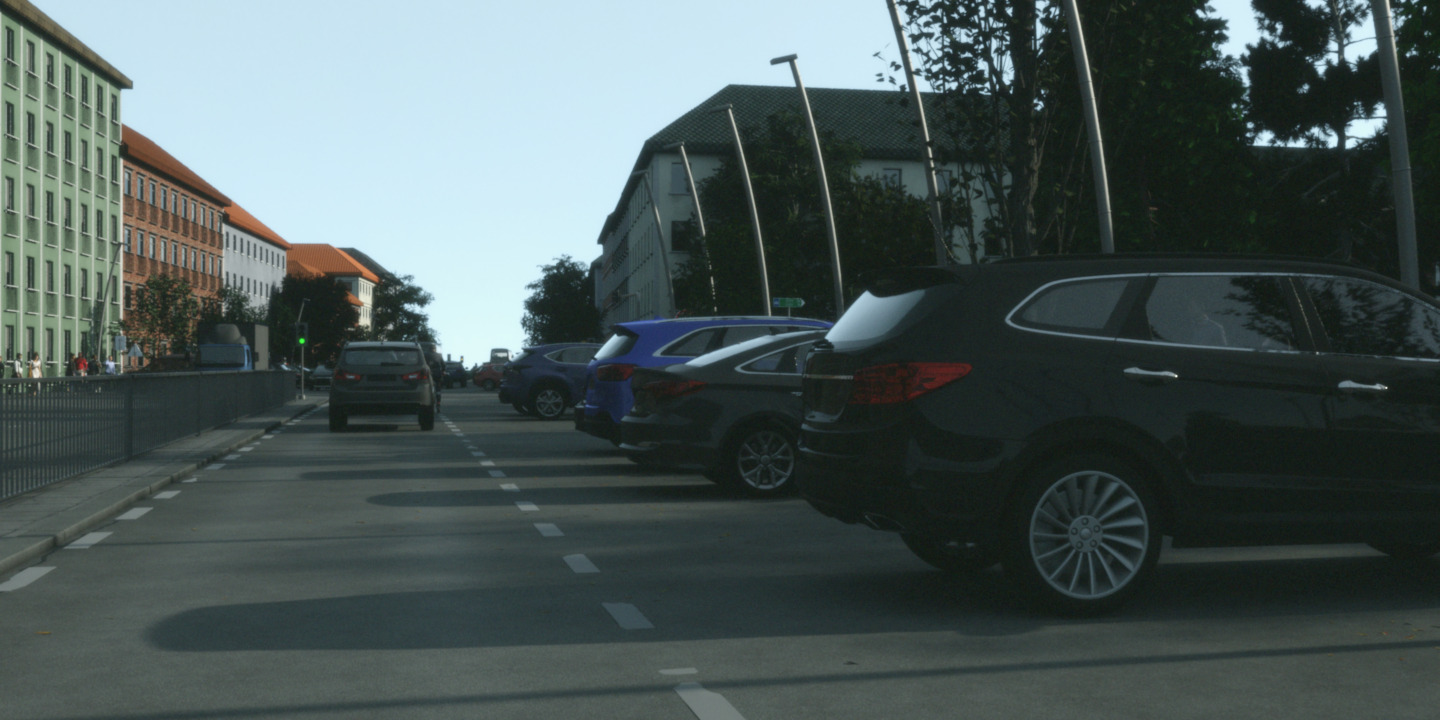}  
		\includegraphics[width=0.32\linewidth]{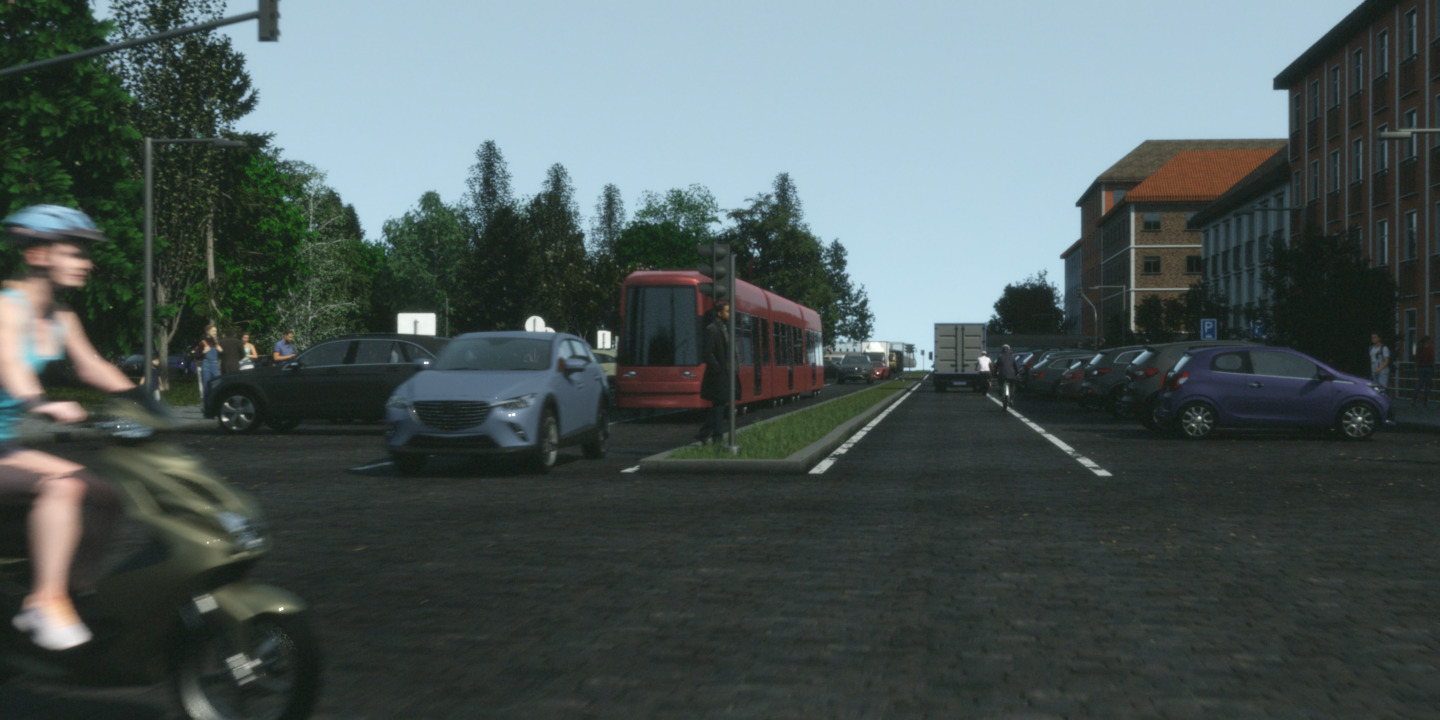}  
		\caption{Example Synthetic images from Synscapes.}
		\label{fig:synscape_input_3}
	\end{subfigure}   
	
	\begin{subfigure}{0.9\linewidth}  
		\centering  
		\includegraphics[width=0.32\linewidth]{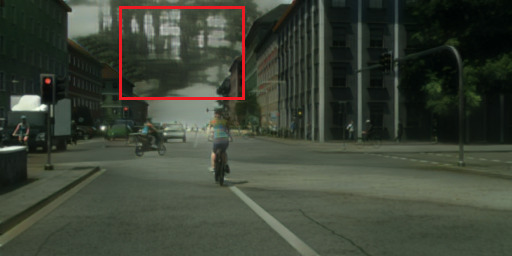}
		\includegraphics[width=0.32\linewidth]{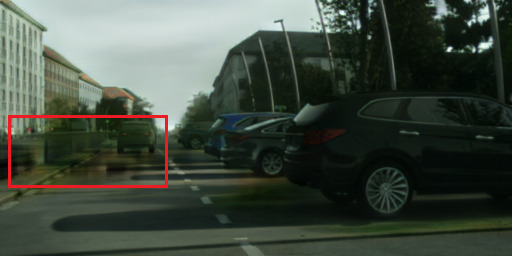}  
		\includegraphics[width=0.32\linewidth]{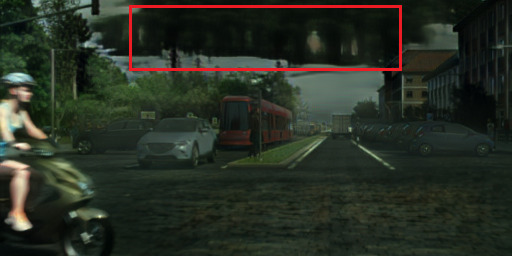}  
		\caption{Results using original UNIT-GAN architecture.}
		\label{fig:unit_gan_output_3}
	\end{subfigure}
	
	\begin{subfigure}{0.9\linewidth}  
		\centering  
		\includegraphics[width=0.32\linewidth]{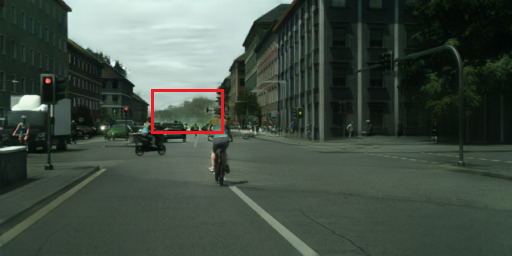}
		\includegraphics[width=0.32\linewidth]{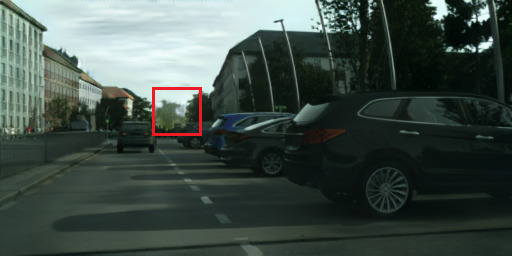}  
		\includegraphics[width=0.32\linewidth]{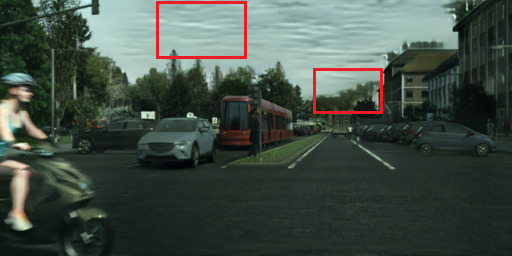}  
		\caption{Results using SP-Feat based Batch selection strategy.}
		\label{fig:spfeat_output_3}
	\end{subfigure}
	
	\caption{Perturbations in transformed images using original UNIT-GAN and our proposed method. \textit{Red boxes} highlight noises in (c) are semantically coherent and less arbitrary, when compared to the noise from original UNIT-GAN network (b). Best viewed in color and by zooming in.}
	\label{fig:qualitative_result_3}
\end{figure*}

In figure \ref{fig:qualitative_result_3} we show some of the failure cases visually for some of the transformed images using our proposed approach.  However, we would like to point out one interesting observation from the presented images. Both basic UNIT-GAN and our modified approach produce certain perceptually inferior quality images. But on closer inspection, we can observe a subtle difference in the \textit{noise} aspect between them. Unlike figure \ref{fig:unit_gan_output} and \ref{fig:unit_gan_output_3}, the perturbations in figure \ref{fig:spfeat_output_3} appear to be less `random' and bears some semantic relevance to what we might expect in similar  scenarios. For instance, it is not uncommon to come across images containing vegetation at the end of the road, or creates cloud like structure in the sky. In fact, in Cityscapes we can find similar layout in a lot of training images. This further reinforces the motivation to use semantically consistent pairs from synthetic and actual data in our \textit{smart batch-loader} approach. It enables the generator to better capture the semantic mapping between the two domains, even though the supervision of such semantic labels is not integrated to its training. The added benefit of providing semantic consistency as a pre-processing to the batch sampling strategy is two-folds: a) it can be easily adopted to train any GAN architecture and b) once training is complete, the network no longer requires it when used for inferencing. 

\section{Conclusion}

In this paper, we proposed a simple and novel, semantic layout based image pair selection strategy and evaluated its advantage for reducing domain discrepancy between the features of synthetic and real world images. The method makes use of semantically consistent image pairs only during training which can be easily obtained using relevant meta-information. This method can be seen as the solution to a potential problem in image-to-image translation tasks: semantic inconsistencies across the batches of source domain and target domain. It can be used in association with any domain translation model to improve the adaptability between the two domains. 

In autonomous driving, for instance, semantic segmentation plays a vital role and our experiments showed the benefit of our approach in such a task. Extensive evaluations are carried out on the challenging Synscapes dataset, showcasing the strength of our proposed contributions. Our method provided considerable gains on FID scores besides improving the quality of transformed images.

In future, we would like to derive semantic resemblance between the synthetic and real world images from the raw images themselves. That is the extracted feature should work across domains, agnostic to the underlying distribution difference. Our proposed SP-Feat based approach achieves this using pixel-level semantic information.

\bibliographystyle{unsrt}  
\bibliography{templateArxiv}  

\begin{thebibliography}{10}

\bibitem{unitgan}
Ming{-}Yu Liu, Thomas Breuel, and Jan Kautz.
\newblock Unsupervised image-to-image translation networks.
\newblock {\em Advances in Neural Information Processing Systems (NeurIPS)},
  2017.

\bibitem{cityscapes}
Marius Cordts, Mohamed Omran, Sebastian Ramos, Timo Rehfeld, Markus Enzweiler,
  Rodrigo Benenson, Uwe Franke, Stefan Roth, and Bernt Schiele.
\newblock The cityscapes dataset for semantic urban scene understanding.
\newblock In {\em Proc. of the IEEE Conference on Computer Vision and Pattern
  Recognition (CVPR)}, 2016.

\bibitem{synscapes}
Magnus Wrenninge and Jonas Unger.
\newblock Synscapes: {A} photorealistic synthetic dataset for street scene
  parsing.
\newblock {\em CoRR}, abs/1810.08705, 2018.

\bibitem{admarket}
BusinessWire.
\newblock Autonomous vehicle market outlook - 2026, 2019.
\newblock
  \url{https://www.businesswire.com/news/home/20190719005370/en/Global-Autonomous-Vehicles-Market-Outlook-2026--},
  Sidst set 14/11/2019.

\bibitem{waymo}
The Verge.
\newblock Waymo’s autonomous cars have driven 8 million miles on public
  roads, 2018.
\newblock
  \url{https://www.theverge.com/2018/7/20/17595968/waymo-self-driving-cars-8-million-miles-testing},
  Sidst set 14/11/2019.

\bibitem{fcn}
Judy Hoffman, Dequan Wang, Fisher Yu, and Trevor Darrell.
\newblock Fcns in the wild: Pixel-level adversarial and constraint-based
  adaptation.
\newblock {\em ArXiv}, abs/1612.02649, 2016.

\bibitem{gan}
Ian Goodfellow, Jean Pouget-Abadie, Mehdi Mirza, Bing Xu, David Warde-Farley,
  Sherjil Ozair, Aaron Courville, and Yoshua Bengio.
\newblock Generative adversarial nets.
\newblock In {\em Advances in Neural Information Processing Systems 27}, pages
  2672--2680. 2014.

\bibitem{closed_form_soln}
Yijun Li, Ming{-}Yu Liu, Xueting Li, Ming{-}Hsuan Yang, and Jan Kautz.
\newblock A closed-form solution to photorealistic image stylization.
\newblock {\em CoRR}, abs/1802.06474, 2018.

\bibitem{gtav}
Stephan~R. Richter, Vibhav Vineet, Stefan Roth, and Vladlen Koltun.
\newblock Playing for data: {G}round truth from computer games.
\newblock In {\em European Conference on Computer Vision (ECCV)}, volume 9906
  of {\em LNCS}, pages 102--118. Springer International Publishing, 2016.

\bibitem{image_analogies}
Aaron Hertzmann, Charles~E. Jacobs, Nuria Oliver, Brian Curless, and David~H.
  Salesin.
\newblock Image analogies.
\newblock In {\em Proceedings of the 28th Annual Conference on Computer
  Graphics and Interactive Techniques}, SIGGRAPH '01, pages 327--340, New York,
  NY, USA, 2001. ACM.

\bibitem{texture_synthesis}
Alexei~A. Efros and Thomas~K. Leung.
\newblock Texture synthesis by non-parametric sampling.
\newblock In {\em Proceedings of the International Conference on Computer
  Vision-Volume 2 - Volume 2}, ICCV '99, pages 1033--, Washington, DC, USA,
  1999. IEEE Computer Society.

\bibitem{vae}
Diederik~P. Kingma and Max Welling.
\newblock Auto-encoding variational bayes.
\newblock {\em CoRR}, abs/1312.6114, 2013.

\bibitem{attention_model}
Karol Gregor, Ivo Danihelka, Alex Graves, and Daan Wierstra.
\newblock {DRAW:} {A} recurrent neural network for image generation.
\newblock {\em CoRR}, abs/1502.04623, 2015.

\bibitem{moment_matching}
Yujia Li, Kevin Swersky, and Richard Zemel.
\newblock Generative moment matching networks.
\newblock In {\em Proceedings of the 32Nd International Conference on
  International Conference on Machine Learning - Volume 37}, ICML'15, pages
  1718--1727. JMLR.org, 2015.

\bibitem{stochastic_backprop}
Danilo~Jimenez Rezende, Shakir Mohamed, and Daan Wierstra.
\newblock Stochastic backpropagation and approximate inference in deep
  generative models, 2014.

\bibitem{diffusion_process}
Jascha Sohl{-}Dickstein, Eric~A. Weiss, Niru Maheswaranathan, and Surya
  Ganguli.
\newblock Deep unsupervised learning using nonequilibrium thermodynamics.
\newblock {\em CoRR}, abs/1503.03585, 2015.

\bibitem{pix2pix}
Phillip Isola, Jun{-}Yan Zhu, Tinghui Zhou, and Alexei~A. Efros.
\newblock Image-to-image translation with conditional adversarial networks.
\newblock {\em CoRR}, abs/1611.07004, 2016.

\bibitem{sketch_learning}
Patsorn Sangkloy, Jingwan Lu, Chen Fang, Fisher Yu, and James Hays.
\newblock Scribbler: Controlling deep image synthesis with sketch and color.
\newblock {\em CoRR}, abs/1612.00835, 2016.

\bibitem{attribute_learning}
Levent Karacan, Zeynep Akata, Aykut Erdem, and Erkut Erdem.
\newblock Learning to generate images of outdoor scenes from attributes and
  semantic layouts.
\newblock {\em CoRR}, abs/1612.00215, 2016.

\bibitem{deep_style_transfer}
Fujun Luan, Sylvain Paris, Eli Shechtman, and Kavita Bala.
\newblock Deep photo style transfer.
\newblock {\em CoRR}, abs/1703.07511, 2017.

\bibitem{learning_to_generate}
Levent Karacan, Zeynep Akata, Aykut Erdem, and Erkut Erdem.
\newblock Learning to generate images of outdoor scenes from attributes and
  semantic layouts.
\newblock {\em CoRR}, abs/1612.00215, 2016.

\bibitem{spade}
Taesung Park, Ming-Yu Liu, Ting-Chun Wang, and Jun-Yan Zhu.
\newblock Semantic image synthesis with spatially-adaptive normalization.
\newblock In {\em Proceedings of the IEEE Conference on Computer Vision and
  Pattern Recognition}, 2019.

\bibitem{cycada}
Judy Hoffman, Eric Tzeng, Taesung Park, Jun{-}Yan Zhu, Phillip Isola, Kate
  Saenko, Alexei~A. Efros, and Trevor Darrell.
\newblock Cycada: Cycle-consistent adversarial domain adaptation.
\newblock {\em CoRR}, abs/1711.03213, 2017.

\bibitem{infogan}
Xi~Chen, Yan Duan, Rein Houthooft, John Schulman, Ilya Sutskever, and Pieter
  Abbeel.
\newblock Infogan: Interpretable representation learning by information
  maximizing generative adversarial nets.
\newblock {\em CoRR}, abs/1606.03657, 2016.

\bibitem{discogan}
Taeksoo Kim, Moonsu Cha, Hyunsoo Kim, Jung~Kwon Lee, and Jiwon Kim.
\newblock Learning to discover cross-domain relations with generative
  adversarial networks.
\newblock {\em CoRR}, abs/1703.05192, 2017.

\bibitem{diverse_translation}
Hsin{-}Ying Lee, Hung{-}Yu Tseng, Jia{-}Bin Huang, Maneesh~Kumar Singh, and
  Ming{-}Hsuan Yang.
\newblock Diverse image-to-image translation via disentangled representations.
\newblock {\em CoRR}, abs/1808.00948, 2018.

\bibitem{DRIT_plus}
Hsin-Ying Lee, Hung-Yu Tseng, Qi~Mao, Jia-Bin Huang, Yu-Ding Lu, Maneesh~Kumar
  Singh, and Ming-Hsuan Yang.
\newblock Drit++: Diverse image-to-image translation viadisentangled
  representations.
\newblock {\em arXiv preprint arXiv:1905.01270}, 2019.

\bibitem{cogan}
Ming{-}Yu Liu and Oncel Tuzel.
\newblock Coupled generative adversarial networks.
\newblock {\em CoRR}, abs/1606.07536, 2016.

\bibitem{cross_modal}
Yusuf Aytar, Llu{\'{\i}}s Castrej{\'{o}}n, Carl Vondrick, Hamed Pirsiavash, and
  Antonio Torralba.
\newblock Cross-modal scene networks.
\newblock {\em CoRR}, abs/1610.09003, 2016.

\bibitem{cyclegan}
Jun{-}Yan Zhu, Taesung Park, Phillip Isola, and Alexei~A. Efros.
\newblock Unpaired image-to-image translation using cycle-consistent
  adversarial networks.
\newblock {\em CoRR}, abs/1703.10593, 2017.

\bibitem{munit}
Xun Huang, Ming-Yu Liu, Serge Belongie, and Jan Kautz.
\newblock Multimodal unsupervised image-to-image translation.
\newblock In {\em ECCV}, 2018.

\bibitem{erfnet}
Eduardo Romera, Jos{\'e}~Manuel {\'A}lvarez, Luis~Miguel Bergasa, and Roberto
  Arroyo.
\newblock Erfnet: Efficient residual factorized convnet for real-time semantic
  segmentation.
\newblock {\em IEEE Transactions on Intelligent Transportation Systems},
  19:263--272, 2018.

\bibitem{imagenet}
J.~Deng, W.~Dong, R.~Socher, L.-J. Li, K.~Li, and L.~Fei-Fei.
\newblock {ImageNet: A Large-Scale Hierarchical Image Database}.
\newblock In {\em Proc. of the IEEE Conference on Computer Vision and Pattern
  Recognition (CVPR)}, 2009.

\bibitem{inceptionv3}
Christian Szegedy, Vincent Vanhoucke, Sergey Ioffe, Jon Shlens, and ZB~Wojna.
\newblock Rethinking the inception architecture for computer vision.
\newblock In {\em Proc. of the IEEE Conference on Computer Vision and Pattern
  Recognition (CVPR)}, 2016.

\end{thebibliography}

\end{document}